\documentclass[a4paper]{article}

\usepackage[english]{babel}
\usepackage[utf8x]{inputenc}
\usepackage[T1]{fontenc}

\usepackage[a4paper,top=3cm,bottom=2cm,left=3cm,right=3cm,marginparwidth=1.75cm]{geometry}

\usepackage{amsmath}
\usepackage{amssymb}
\usepackage{graphicx}
\usepackage[colorinlistoftodos]{todonotes}
\usepackage{hyperref}
\hypersetup{colorlinks=false, allcolors=black, citecolor=black}

\usepackage[export]{adjustbox}
\usepackage[boxed]{algorithm2e}
\usepackage{color}
\usepackage{colortbl}
\usepackage{comment}
\usepackage[outline]{contour}
\usepackage{epstopdf}
\usepackage{fancyvrb}
\usepackage{filecontents}
\usepackage{float}
\usepackage[stable]{footmisc}
\usepackage{gensymb}
\usepackage{graphicx}
\usepackage{mathtools}
\usepackage{multicol}
\usepackage{multirow}
\usepackage[numbers]{natbib}
\usepackage{tcolorbox}
\usepackage{titlesec}
\usepackage{dsfont}
\usepackage{soul}

\titleformat{\paragraph}
{\normalfont\normalsize\bfseries}{\theparagraph}{1em}{}
\titlespacing*{\paragraph}
{0pt}{3.25ex plus 1ex minus .2ex}{1.5ex plus .2ex}

\definecolor{CodeGray}{RGB}{240,240,240}

\newcommand{\inlinecode}[1]{\colorbox{CodeGray}{\texttt{#1}}}

\newcommand{\clrBlue}[1]{{\textcolor[HTML]{0074D9}{#1}}}
\newcommand{\clrOlive}[1]{{\textcolor[HTML]{3D9970}{#1}}}
\newcommand{\clrFuchsia}[1]{{\textcolor[HTML]{F012BE}{#1}}}

\newcommand{\xdownarrow}[1]{%
  {\left\downarrow\vbox to #1{}\right.\kern-\nulldelimiterspace}
}

\begin{document}

\title{Optimal to-do list gamification}

\author{
    Jugoslav Stojcheski, Valkyrie Felso, \& Falk Lieder\\
    Rationality Enhancement Group \\
	Max Planck Institute for Intelligent Systems \\
    Tübingen, Baden-Württemberg, Germany
}

\date{\today}

\vspace{0.2in}
\maketitle

\begin{abstract}

What should I work on first? What can wait until later? Which projects should I prioritize and which tasks are not worth my time? These are challenging questions that many people face every day. People's intuitive strategy is to prioritize their immediate experience over the long-term consequences. This leads to procrastination and the neglect of important long-term projects in favor of seemingly urgent tasks that are less important. Optimal gamification strives to help people overcome these problems by incentivizing each task by a number of points that communicates how valuable it is in the long-run. Unfortunately, computing the optimal number of points with standard dynamic programming methods quickly becomes intractable as the number of a person's projects and the number of tasks required by each project increase. Here, we introduce and evaluate a scalable method for identifying which tasks are most important in the long run and incentivizing each task according to its long-term value. Our method makes it possible to create to-do list gamification apps that can handle the size and complexity of people's to-do lists in the real world.\\


\end{abstract}
\newpage

\tableofcontents
\newpage

\section{Introduction}  \label{sec:intro}

    % ===== Big picture =====
    Many people struggle with procrastination and setting priorities \citep{Steel2007}. 
    But prioritization is difficult and self-discipline is challenging. Decision support systems have been developed to support human decision-making in very specific domains \cite{aviv2005partially, bhatnagar1999markov, gadomski2001towards, nunes2009markov, song2000optimal}.  But, to date, there is no decision-support system for deciding what to work on.
    
    % ===== Specific issue =====
    Procrastination takes a considerable toll on people’s lives, the economy and society at large. It is often a consequence of people’s propensity to prioritize their immediate experiences over the long-term consequences of their actions \citep{Steel2007}. This is problematic when immediate rewards are misaligned with long-term value. This can be addressed by optimal to-do-list gamification \cite{lieder2019cognitive} which proposes adding incentives to tasks to help people overcome procrastination. However, so far, this approach has been limited to small artificial laboratory paradigms in which people's to-do lists were restricted due to the exponentially-increasing computational cost with respect to the number of tasks     in the to-do list. Mitigating this problem so that to-do list gamification can be applied to real-world to-do lists was the major issue to be addressed.
    
    % ===== Approach | Here we describe the key idea for how to make to-do list gamification scalable. =====
    In this text, we present an algorithmic solution that tackles the scalability issue.    The key idea is to leverage the natural structure of a to-do list composed of goals and tasks, and to use a 2-level hierarchical decomposition of a discrete-time semi-Markov decision process (SMDP; \cite{howard1963semi}). In addition, we introduce inductive biases in the procedure of searching for an optimal solution, which removes a large amount of unnecessary computations.
    We show empirically that this method is drastically more scalable than other methods used previously for tackling this problem such as \textit{Backward Induction} \cite{zermelo1913anwendung} and \textit{Value Iteration} \cite{bellman1957markovian}. Concretely, we provide results
    in terms of wall-clock computational time for to-do lists with varying numbers of goals and tasks.

    % ===== Payoffs | What are the benefits of this approach? =====
    The main benefit of introducing hierarchies in the procedure of solving SMDPs is the large reduction of the state-space complexity. Moreover, the introduced inductive biases reduce the the problem complexity by discarding sequences of actions that cannot possibly be part of an optimal solution. These improvements lead to both a massive computational speed-up and low memory requirements. This enables the gamification application to process large to-do lists that were previously intractable, which in turn allows users to make detailed plans for multiple goals far into the future. We expect the performance of the algorithm presented here to be further improved in future work by adding more layers of abstraction in the hierarchy of SMDP solvers and by introducing additional inductive biases.

    Furthermore, we developed an application programming interface (API) that takes in a to-do list and outputs a gamified list of tasks for a desired workload. However, although there are no known limitations to the algorithmic procedure, our API is currently unable to process some large to-do lists due to a 30-second request timeout. We are confident that future work will be able to address this issue by changing the software infrastructure. We provide details on this limitation in Section~\ref{sec:evalAPI}.
        
    % ===== Outline | Outline of the remainder of the report. =====
    This text is divided into the following sections:
    In Section~\ref{sec:problem_def}, we describe the problem that we are trying to solve, and we present a formal definition of the problem as a discrete-time semi-Markov decision process.
    In Section~\ref{sec:solution}, we describe two algorithmic tricks that allow us to solve this problem more efficiently, the API components, their functionalities, and the way solutions are generated.
    In Section~\ref{sec:api}, we give a brief introduction to the API that we have developed. 
    In Section~\ref{sec:evaluations}, we provide a use case in which we inspect pseudo-rewards, evaluate the speed and reliability of the API, and make a qualitative comparison between non-hierarchical and hierarchical SMDP solving methods.
    Lastly, in Section~\ref{sec:future_work} we outline directions for future work, and the Appendix provides details on the API's input and output.

\section{Problem definition}  \label{sec:problem_def}
    The goal of to-do list gamification is to  maximize the user's long-term productivity\footnote{We define \textit{productivity} as the amount of \textit{value} that a person generates by completing a series of tasks within a fixed amount of time.} by proposing daily task lists where each task is incentivized by a certain number of points.
    Users compose hierarchical \textbf{to-do lists} comprised of two types of items: \textit{goals} and \textit{tasks}. Each \textbf{goal} contains a deadline, a value estimate, and a list of tasks which contribute to the goal. Each \textbf{task} contains a \textit{time estimate}. Tasks may optionally include a \textit{deadline} and \textit{scheduling tags} (i.e. ``do dates" and ``do days").
    In addition, users provide their desired workload in hours for a typical day (\textit{typical day's working hours}) and for the day at hand (\textit{today's working hours}). 
    The outputted gamified daily schedule should contain all tasks that users indicated they wanted to work on today, as well as additional tasks towards their goals, up to the desired daily workload. Further details on these components of the to-do list are provided in the Appendix.

    We formally define this problem in terms of a discrete-time semi-Markov decision process (SMDP) in the sections that follow. We first model working on projects as an SMDP in Section~\ref{sec:smdp}. Then, we build on this model to formalize the problem of productivity maximization by optimal gamification in Section~\ref{sec:optimalGamification}. 

    \subsection{Modeling working on projects as a discrete-time semi-Markov decision process}  \label{sec:smdp}
    
        The definition of a discrete-time semi-Markov decision process (SMDP) $M = (\mathcal{S}, \mathcal{A}, F, T, r, \gamma)$ comprises state space $\mathcal{S}$, action space $\mathcal{A}$, transition-time function $F$, transition dynamics $T$, reward function $r$, and a discount factor $\gamma$.
        We model each level in a hierarchical to-do list as an SMDP.
        For simplicity, we initially concentrate on a 2-level hierarchical decomposition, in which we have one goal-level SMDP ($M_{\text{(goal)}}$) and one task-level SMDP for each goal ($M_{\text{(task)}}^{(g)}$).
        Since tasks which have to be completed in the same period of time or which contribute to a sub-goal can be grouped to form an SMDP on their own, adding intermediate layers of abstraction between the goal level and the task level is also viable but is left for future work.
        
        We provide a detailed description of these parameters in the following sections. Additionally, we give a formal definition of the optimal policy in relation to the traditional definition of Q-function in Section~\ref{sec:policy}.

        \subsubsection{State space $\mathcal{S}$}  \label{smdp:state_space}
        
            As mentioned in the introductory section (Section~\ref{sec:intro}), the main benefit of the hierarchical decomposition of the problem is the huge reduction of the state-space cardinality. In this section, we present an informal complexity analysis of this improvement by first defining the non-hierarchical state space of the problem, and then defining the hierarchical state space of the problem.

            The state space of the original (non-hierarchical) SMDP - $\mathcal{S}$ - consists of all possible combinations of completed and uncompleted tasks. 
            Intuitively, since a task in a to-do list can be either completed or not completed, we represent each instance $s \in \mathcal{S}$ of the state space as a binary vector whose length corresponds to the total number of tasks in a to-do list. For example, if the total number of tasks is $n$, then the state space $\mathcal{S}$ is defined by an $n$-fold Cartesian product of the set $\{ 0, 1 \}$. Formally written, that is
            $$
                \mathcal{S} = \left\{ \{ 0, 1 \}^{n} \right\}
            $$
            where the $i$-th element of any binary vector $s \in \mathcal{S}$ is $1$ if the task associated with that element is completed and $0$ otherwise.

            According to this definition, the state space comprises all possible configurations of the binary vector and its complexity is exponential with respect to the total number of tasks $n$. Formally, the cardinality of the state space is exactly $2^{n}$.
            For a real-world to-do list with a non-trivial number of tasks, this is an extremely large number of states, and operating on it demands enormous amounts of computational power and memory.
            
            In order to reduce the cardinality of the state space, we introduce a 2-level hierarchical decomposition that decomposes the state space into mutually exclusive sets:
            \begin{enumerate}
                \item \textbf{Goal-level state space} $\mathcal{S}_{\text{(goal)}}$, in which we keep track of goals' completion.
                \item \textbf{Task-level state space} $\mathcal{S}_{\text{(task)}}$ for each goal, in which we keep track of tasks' completion within a goal.
            \end{enumerate}
            
            The goal-level state space $\mathcal{S}_{\text{(goal)}}$ consists of a binary vector of length $|\mathbf{\mathcal{G}}|$, where $\mathbf{\mathcal{G}}$ is the set of goals and $|\mathbf{\mathcal{G}}|$ is the total number of goals in a to-do list.
            $$
                \mathcal{S}_{\text{(goal)}} = \left\{ \{ 0, 1 \}^{|\mathbf{\mathcal{G}}|} \right\}
            $$
            
            In a similar manner, we define the task-level state space $\mathcal{S}_{\text{(task)}}$ for each goal. A task-level state space for an arbitrary goal $g \in \mathcal{G}$ consists of a binary vector of length $n_{g}$, where $n_{g}$ is the total number of tasks associated with that goal.
            $$
                \mathcal{S}_{\text{(task)}}^{(g)} = \left\{ \{ 0, 1 \}^{n_{g}} \right\}
            $$

            On each level, there is only one initial state, i.e. the binary vector of completed and uncompleted goals/tasks at present time.
            If there are no completed goals/tasks at that time, then the initial state is represented by a vector of all zeros - $\mathbf{0}$, i.e.
            $$
                (\text{completed tasks}) = \mathbf{0} = (0, 0, \ldots, 0)
            $$
            Furthermore, all states in which all goals/tasks are completed are considered to be terminal states as well as states that follow as a consequence of taking a special kind of action (i.e. \textit{slack-off action}, see Section~\ref{sec:action_space}). In the former case, the terminal state is represented by a vector of all ones - $\mathbf{1}$, i.e.
            $$
                (\text{completed tasks}) = \mathbf{1} = (1, 1, \ldots, 1)
            $$
            In order to avoid that taking the last step toward achieving a goal could be penalized by a negative pseudo-reward (see Equation~\ref{eq:pseudoRewards}), we introduce an additional state, a so-called \textit{goal-achieving state} - $s_{\dagger}$. The purpose of this special state is to separate the cost of action execution from the reward for goal-achievement reward. The goal-achieving state can only be reached as a consequence of \textit{automatic} execution of an instantaneous action $a_{\dagger}$ immediately after reaching the terminal state in which all tasks are completed (i.e. $\mathbf{1}$).

            We do a non-rigorous complexity analysis in order to show the state-space reduction obtained as a consequence of the hierarchical decomposition in the following manner.
            Let $n_{g}$ be the total number of tasks associated with goal $g \in \mathcal{G}$. Assuming mutually-exclusive task-level state spaces, the total number of tasks in a to-do list - $n$ - can be written as
            $$
                n = \sum_{g=1}^{|\mathbf{\mathcal{G}}|} n_{g}
            $$
            Motivated by real-world to-do lists, we assume that each to-do list is composed of at least two goals (i.e. $|\mathbf{\mathcal{G}}| \ge 2$) and each goal has at least two tasks (i.e. $n_{g} \ge 2$, $g = 1, \ldots, |\mathbf{\mathcal{G}}| $) associated with it. In this case, the following strict inequality holds:
            $$
                | \mathcal{S} |
                = 2^{n} 
                = 2^{\sum_{g=1}^{|\mathbf{\mathcal{G}}|} n_{g}}
                = \prod_{g=1}^{|\mathbf{\mathcal{G}}|} 2^{n_{g}}
                > 2^{|\mathbf{\mathcal{G}}|} + \sum_{g=1}^{|\mathbf{\mathcal{G}}|} 2^{n_{g}}
                = | \mathcal{S}_{\text{(goal)}} | + | \mathcal{S}_{\text{(task)}} |
            $$
            Moreover, the exact reduction of the state-space cardinality (as a percentage) can be computed as
            $$
                \left( 1 - \dfrac
                    { 2^{|\mathbf{\mathcal{G}}|} + \sum_{g=1}^{|\mathbf{\mathcal{G}}|} 2^{n_{g}} }
                    { \prod_{g=1}^{|\mathbf{\mathcal{G}}|} 2^{n_{g}} } \right) \cdot 100 \%
            $$

        \subsubsection{Action space $\mathcal{A}$}  \label{sec:action_space}
        
            In a similar manner to the state space decomposition, we decompose the action space into two mutually-exclusive sets, a goal-level action space 
            ($\mathcal{A}_{\text{(goal)}}$) and task-level action space ($\mathcal{A}_{\text{(task)}}^{(g)}$) for each goal $g \in \mathcal{G}$.

            The goal-level action space $\mathcal{A}_{\text{(goal)}}$ consists of all goals in a to-do list as well as one \textit{slack-off} action $a_{+}$.
            $$
                \mathcal{A}_{\text{(goal)}} = \left\{ a_{g} \ | \ g = 1, \ldots, |\mathbf{\mathcal{G}}| \right\} \cup \{ a_{+} \}
            $$
            
            The task-level action space $\mathcal{A}_{\text{(task)}}^{(g)}$ for a goal $g$ consists of all tasks within that goal as well as one \textit{slack-off} action $a_{+}$.
            $$
                \mathcal{A}_{\text{(task)}}^{(g)} = \left\{ a_{i}^{(g)} \ | \ i = 1, \ldots, n_{g} \right\} \cup \{ a_{+}^{(g)} \}  
            $$
            
            A \textit{slack-off} action is a special kind of action whose purpose is to model the value of an immediately-enjoyable action by providing an immediate positive reward after its execution.
            Alternatively, it can be used by the policy to discard goals that are not worth attaining.
            However, it is not explicitly a part of the to-do list, and the \textit{slack-off} action is therefore \textit{not} a part of the state space on any level.
            This special kind of action can be executed at any time and its initial execution triggers an infinite sequence of \textit{slack-off}-action executions since there exists no other more-valuable action.
            At this point, the value of this action is arbitrarily chosen to be a small positive constant, but future work will quantify how much people value their leisure time relative to the value of attaining their goals. Moreover, the value of a \textit{slack-off} action might be dependent on the level at which it occurs and/or the goal that it is associated with.

            Any goal-level action can be chosen and executed as long as there are uncompleted tasks associated with that goal. Alternatively, the \textit{slack-off} action can be chosen. The set of all \textit{available} goal-level actions is completely determined by the \textit{history} of task-level actions for each goal. Formally, the availability of the goal-level action $a_{g}$ is directly associated with the value of the $g$-th entry in the binary vector $s$, i.e.
            $$
                s_{g} = \left( \bigwedge\limits_{i=1}^{n_{g}} a_{i}^{(g)} \right) \vee a_{+}^{(g)}
            $$
            where the goal-state action $a_{g}$ is available if $s_{g}$ is false and is unavailable otherwise. Please note that in order to simplify the notation, in the rest of this text, we will use $s_{t}$ to denote a goal-level state that occurs at \textit{time} $t \in \mathbb{Z}_{\ge 0}$, and $\tilde{s}_{t}^{(g)}$ to denote a task-level state associated with goal $g$ that occurs at \textit{time} $t \in \mathbb{Z}_{\ge 0}$.

            Under the assumption that users \textbf{always} complete a task successfully once they start working on it, we allow each task-level action
            (except for the \textit{slack-off} action)
            to be performed \textbf{at most once}.
            Therefore, for a goal $g$, the set of all \textit{available} task-level actions (i.e. uncompleted tasks) at time $\tilde{t}$ $\left(\Omega_{\tilde{t}}^{(g)} \subseteq \mathcal{A}_{\text{(task)}}^{(g)}\right)$ depends on the \textit{history} of actions (i.e. completed tasks) by that time $\left(H_{\tilde{t}}^{(g)} \subseteq \mathcal{A}_{\text{(task)}}^{(g)}\right)$. Using set notation, we can formally write this as:
            $$
                \Omega_{\tilde{t}}^{(g)} = \{ a \ | \ a \in \mathcal{A}_{\text{(task)}}^{(g)} \land a \notin H_{\tilde{t}}^{(g)} \} \cup \{ a_{+}^{(g)} \}
                \hspace{0.5cm} \Omega_{\tilde{t}}^{(g)} \cup H_{\tilde{t}}^{(g)} = \mathcal{A}_{\text{(task)}}^{(g)}
                \hspace{0.5cm} \Omega_{\tilde{t}}^{(g)} \cap H_{\tilde{t}}^{(g)} = \{ a_{+}^{(g)} \}
            $$

            Furthermore, we consider task-level action execution to be indivisible. This is analogous to a user working on only one task at a time until that task is completely executed. In other words, the agent is allowed to choose an action to perform next only if there is no other action in progress at that time. After executing an action $a$, the action is marked as completed by changing the value of the corresponding element in the task-level binary state vector from $0$ to $1$.
            As a consequence, the SMDP moves $\tilde{\tau}$ time units in the future according to the task time estimate (e.g. number of minutes), which is stochastically obtained by the transition-time function $F$ (defined in Section~\ref{sec:smdp_time}). Additionally, if the executed action is \textit{not} the \textit{slack-off} action, we remove the task-level action from the set of available task-level actions $\Omega^{(g)}$ and we add it to the history of actions $\mathcal{H}^{(g)}$. That is, using formal set notation, we perform the following updates
            $$
                \Omega_{\tilde{t} + \tilde{\tau}}^{(g)} = \Omega_{\tilde{t}}^{(g)} \textbackslash \{ a \}
                \hspace{0.5cm}  H_{\tilde{t} + \tilde{\tau}}^{(g)} = H_{\tilde{t}}^{(g)} \cup \{ a \}
            $$
            Otherwise, once the \textit{slack-off action} is chosen, the SMDP reaches a terminal state in which the \textit{slack-off action} is executed infinitely many times.

            Similar set operations apply analogously at the goal-level once a goal is completed. Briefly, we check whether the last completed task completes the goal which it contributes to and we change the corresponding value of the goal-level binary state vector from $0$ to $1$ if that is the case.
            
        \subsubsection{Transition time $F$}  \label{sec:smdp_time}
        
            The formalization of a problem within the framework of an SMDP requires a definition of a transition-time function $F$. Here, $F(\tau | s_{t}, a)$ is the probability that the time at which the agent has to make the next decision occurs in exactly $\tau$ time units, as a consequence of executing action $a$ in state $s$ at time $t$.
            We chose a transition-time function that can model the cognitive bias known as the \textit{planning fallacy}, in which people underestimate the time required to complete a task.
            Kahneman and Tversky \cite{kahneman1977intuitive} describe this bias as such: "\textit{Scientists and writers, for example, are notoriously prone to underestimate the time required to complete a project, even when they have considerable experience of past failures to live up to planned schedules... It frequently occurs even when underestimation of duration or cost is actually penalized}." 

            Since people have unreliable time estimates and our SMDPs are decision processes with discrete time steps, we model the number of time units required for action completion to follow a zero-truncated Poisson probability distribution\footnote{Also known as \textit{conditional Poisson distribution}, \textit{positive Poisson distribution}.} with adjusted mean value and variance. We formally define the transition-time function as
            $$
                F(\tau | s_{t}, a)
                := \text{Poisson}_{> 0}(\tau; \tilde{k} )
                = \dfrac{ \tilde{k}^{\tau} e^{- \tilde{k} } }{ \tau! (1 - e^{- \tilde{k} }) }
            $$
            where $\tilde{k} = c_{\text{pf}} \cdot k$, $k$ is the discrete amount of time units required to complete action $a$ in state $s$ at time $t$, and $c_{\text{pf}} \in \mathbb{R}_{> 0}$ is a planning-fallacy constant that adjusts the distribution parameter.
            In lack of knowledge about the exact value of the planning-fallacy constant ($c_{\text{pf}}$), we follow King and Wilson \cite{king1967subjective} and we initially set its value to $1.39$. Obtaining better estimates for this value based on real-world data is left for future work.

        \subsubsection{Transition dynamics $T$}  \label{sec:smdp_transitions}
        
            In the previous sections, we presented two assumptions related to the task completion: (1) users complete a task with certainty once they start working on it, and (2) the time estimate for task completion is stochastic.
            Under these assumptions, the transition dynamics from a current state $s$ at time $t$ to a next state $s'$ after executing an action $a$ is deterministic in completion, but stochastic in duration. In other words, the transition probability $T(s_{t}, a, s_{t + \tau}')$ is completely dependent on the probability of completing an action in exactly $\tau$ time units, which can be formally written as
            $$
                T(s_{t}, a, s_{t + \tau}')
                = \text{Pr}(s_{t+\tau}' | s_{t}, a) \sim F(\tau | s_{t}, a)
                \hspace{0.4cm} \forall s \in \mathcal{S}_{\text{(task)}}^{(g)}
                \hspace{0.4cm} \forall a \in \Omega_{\text{(task)}}^{(g)}
                \hspace{0.4cm} t \in \mathbb{Z}_{\ge 0}
                \hspace{0.4cm} g = 1, \ldots, |\mathbf{\mathcal{G}}|
            $$
            where $\tau \sim F(\tau | s_{t}, a)$ is the transition-time function that determines the time needed to complete a chosen action $a$ in state $s$ at time $t$, and $s'$ is \textit{the} state that follows as a consequence of executing action $a$ in state $s$. Formally, if an action $a$ is represented by the $i$-th bit of the binary state vector $s$, the binary vector of the next state $s'$ can be written as 
            $s' = e_{i} \vee s$,
            where $e_{i}$ is a one-hot vector with a value of $1$  only at its $i$-th position, and $\vee$ represents the ``or'' operation of two binary vectors. This operation is applicable on both the goal and task levels as long as the procedures of goal and task completion described in Section~\ref{sec:action_space} are being completely followed.
            
            A special case of the transition dynamics at task level occurs after reaching the terminal state ($\mathbf{1}$) in which all tasks have been completed. There, the process transits to a goal-achieving state $s_{\dagger}$ after instantaneously executing the action  $a_{\dagger}$  in $0$ time steps, that is
            $
                T(\mathbf{1}, a_{\dagger}, s_{\dagger})
                = \text{Pr}(s_{\dagger} | \mathbf{1}, a_{\dagger})
                = 1
            $.
            
        \subsubsection{Reward function $r$}  \label{sec:reward_fns}

            Formally, we define the goal-level reward function $r_{\text{(goal)}}(s_{t}, a, s_{t + \tau}')$ from a current goal-level state $s$ at time $t$ to a next goal-level state $s'$ at time $t + \tau$ after performing a goal-level action $a \in \Omega_{\text{(goal)}}$ that takes $\tau$ time units for execution
            to be
            $$
                r_{\text{(goal)}}(s_{t}, a, s_{t + \tau}') =
                \begin{dcases*}
                    R(a_{+}) \cdot (1 - \gamma)^{-1}
                        &  \parbox[c]{8cm}{if the goal-level \textit{slack-off} action was chosen}  \\
                        \sum_{\tilde{t}=0}^{\tau-1} r_{\text{(task)}}^{(g)}(\tilde{s}_{\tilde{t}}, \pi_{\text{(task)}}^{(g)}(\tilde{s}_{\tilde{t}}), \tilde{s}'_{\tilde{t}'}) 
                        &  \parbox[c]{8cm}{if any other goal-level action was chosen}
                \end{dcases*}
            $$
            where $a_{+}$ represents the goal-level \textit{slack-off} action, $\gamma \in (0, 1]$ is a discount factor,
            and $r_{\text{(task)}}^{(g)}(\tilde{s}_{\tilde{t}}, \tilde{a}, \tilde{s}_{\tilde{t}'})$ is a task-level reward function of goal $g$.
            
            The task-level reward function $r_{\text{(task)}}^{(g)}(\tilde{s}_{\tilde{t}}, \tilde{a}, \tilde{s}_{\tilde{t} + \tilde{\tau}}')$ for a goal $g$ from a current task-level state $\tilde{s}$ at time $\tilde{t}$ to a next task-level state $\tilde{s}'$ at time $\tilde{t} + \tilde{\tau}$ after performing a task-level action $\tilde{a} \in \Omega_{\text{(task)}}^{(g)}$ that takes $\tilde{\tau}$ time units for execution is defined as
            $$
                r_{\text{(task)}}^{(g)}(\tilde{s}_{\tilde{t}}, \tilde{a}, \tilde{s}_{\tilde{t} + \tilde{\tau}}') = 
                \begin{dcases*}
                    R(\tilde{a}_{+}^{(g)}) \cdot (1 - \gamma)^{-1}
                    & \parbox[c]{8cm}{if the task-level \textit{slack-off} action associated with goal $g$ was chosen}
                    \\
                    - \lambda^{(g)} \sum_{k=0}^{\tilde{\tau}-1} \gamma^{k} 
                    & \parbox[c]{8cm}{if a task-level action with duration $\tilde{\tau}$ was chosen}  \\
                    R(g) \cdot \Pi(\beta_{g}) 
                    & \parbox[c]{8cm}{if the terminal task-level state has been reached, \\ 
                                      i.e. $\tilde{s}_{\tilde{t}} = \mathbf{1}^{(g)}$, $\tilde{a} = a_{\dagger}$, and $\tilde{s}_{\tilde{t} + \tilde{\tau}}' = s_{\dagger}$}
                \end{dcases*}
            $$
            where $\lambda^{(g)} \in \mathbb{R}_{> 0}$ models the value that reflects the cost of a person's time and mental effort to work on goal $g$, and $\tilde{a}_{+}^{(g)}$ is the \textit{slack-off} action associated with goal $g$.

\begin{figure}[!t]
                \centering
                \includegraphics[width=\textwidth]{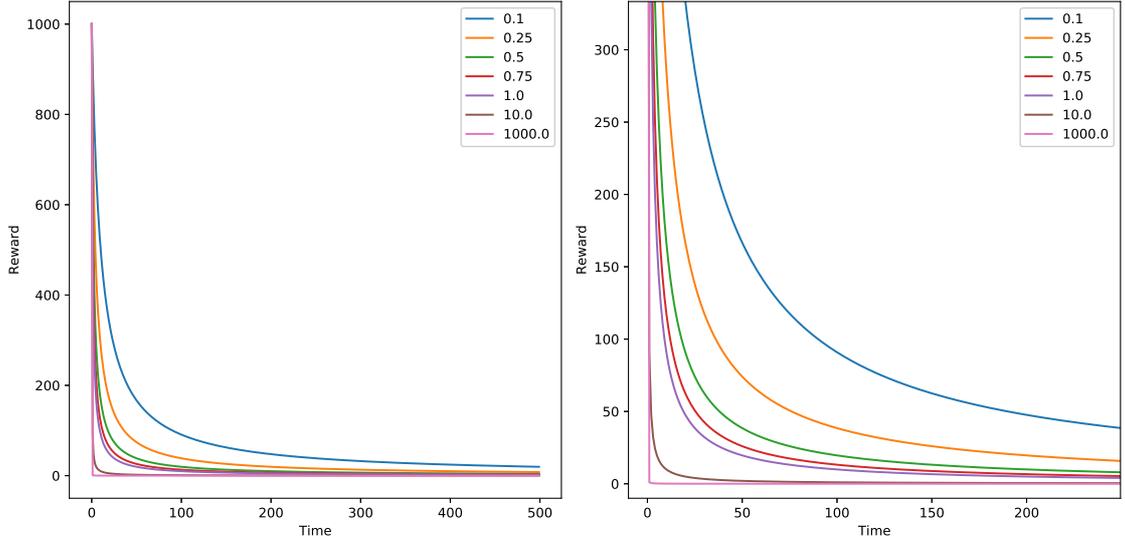}
                \caption{Penalized rewards (vertical axis) for multiple values of $\beta$ (legend) as a function of time (horizontal axis).}
                \label{fig:beta1000}
            \end{figure}

            Furthermore, we define the reward function $R(g)$ as the reward for accomplishing\footnote{A goal is accomplished \textit{iff} each task related to that goal has been completed.} goal $g$. This reward function returns the goal value if executing the next task-level action $\tilde{a}$ leads to completion of the goal.
            In addition, we define $\Pi(\beta_{g})$ to be the penalty function for a goal $g \in \mathcal{G}$.
            The value of the penalty function discounts the goal reward proportionally to the time by which deadlines associated with that goal are missed and it can be formally written as $\Pi(\beta_{g}) = (1 + \beta_{g})^{-1}$.
            Here, $\beta_{g} = \sum_{i=1}^{n_{g}} \psi \cdot \Delta t_{i}$ is a weighted sum of penalties for tasks whose deadlines were missed, where $\psi \in \mathbb{R}_{\ge 0}$ is the penalty rate (per time unit) and $\Delta t_{i}$ is the number of time units by which the deadline was missed. It acts as a constant that regulates the steepness of the curve of goal function $R(g)$. An example of this effect is graphically represented in Figure~\ref{fig:beta1000} for the different values of $\beta$ and a goal value of $1000$.
            
            According to the definition of these reward functions, an immediate negative reward is obtained for completing each task. Conversely, an immediate positive reward is obtained \textit{after} the goal-accomplishing task-level state has been reached or a \textit{slack-off} action has been chosen for execution.
            These definitions of the reward functions are directly related to the inability to see long-term consequences of immediate actions in real life. This poses the problem of misalignment between immediate and long-term rewards which may cause procrastination. We tackle this real-life issue by aligning immediate rewards with future value using reward-shaping functions \cite{ng1999policy} (defined in Section~\ref{sec:optimalGamification}).

        \subsubsection{Optimal policy $\pi^{\star}$}  \label{sec:policy}
        
            First of all, we define the action-value function $Q_{\text{(level)}}$ in a traditional manner. Its definition is exactly the same on both levels, but we explicitly write all expressions for clarity.
            The goal-level action-value function is defined as
            $$
                Q_{\text{(goal)}}(s_{t}, a)
                = \sum_{s' \in \mathcal{S}_{\text{(goal)}}} T(s_{t}, a, s_{t + \tau}')
                  \left[ r_{\text{(goal)}}(s_{t}, a, s_{t + \tau}') 
                         + \gamma^{\tau} \max_{a' \in \Omega_{t + \tau}} Q_{\text{(goal)}}(s_{t + \tau}', a') \right]
            $$
            whereas the task-level action-value function for a goal $g$ is defined as
            $$
                Q_{\text{(task)}}^{(g)}(\tilde{s}_{\tilde{t}}, \tilde{a})
                = \sum_{\tilde{s}' \in \mathcal{S}_{\text{(task)}}^{(g)}}
                  \sum_{\tilde{\tau} \sim F(\cdot | \tilde{s}_{\tilde{t}}, \tilde{a})}
                                  T(\tilde{s}_{\tilde{t}}, \tilde{a}, \tilde{s}_{\tilde{t} + \tilde{\tau}}')
                  \left[ r_{\text{(task)}}^{(g)}(\tilde{s}_{\tilde{t}}, \tilde{a}, \tilde{s}_{\tilde{t} + \tilde{\tau}}') 
                         + \gamma^{\tilde{\tau}} \max_{\tilde{a}' \in \Omega_{\tilde{t} + \tilde{\tau}}^{(g)}} Q_{\text{(task)}}^{(g)}(\tilde{s}_{\tilde{t} + \tilde{\tau}}', \tilde{a}') \right]
            $$
            In both equations, $T(\cdot, \cdot, \cdot)$ is the probability of transition (as defined in Section~\ref{sec:smdp_transitions}), $\tau$ is the time needed to complete a goal-level action $a$, $\tilde{\tau}$ is the time needed to complete a task-level action $\tilde{a}$, and $\gamma \in (0, 1]$ is a discount factor which can be interpreted as the probability that the decision maker can continue to act and gather more rewards when they arrive in state $s'$. Setting $\gamma$ to a value less than $1$ thereby captures the possibility that the episode described by the SMDP can end early so that future rewards might become unavailable.
            
            The optimal policy $\pi_{\text{(level)}}^{\star}$ is an objective function that maximizes the expected sum of discounted future rewards. Here, we extend the definition of an optimal policy to be time-dependent, since the value of the available actions depends not only on the current state $s$ of the SMDP, but also on the time $t$ at which the decision on the next action is made.
            The goal-level optimal policy is defined as
            $$
                \pi_{\text{(goal)}}^{\star}(s_{t_{0}}) = 
                \arg \max_{\pi} \mathds{E} \left[ \sum_{t = t_{0}}^{\infty} \gamma^{t} \cdot r_{\text{(goal)}}(s_{t}, \pi_{\text{(goal)}}(s_{t}), s_{t+\tau}) \right]
            $$
            whereas the task-level optimal policy for a goal $g \in \mathcal{G}$ is defined as
            $$
                \pi_{\text{(task)}}^{(g)\star}(\tilde{s}_{\tilde{t}_{0}}) = \arg \max_{\pi} \mathds{E} \left[ \sum_{\tilde{t} = \tilde{t}_{0}}^{\infty} \gamma^{\tilde{t}} \cdot r_{\text{(task)}}^{(g)}(\tilde{s}_{\tilde{t}}, \pi_{\text{(task)}}^{(g)}(\tilde{s}_{\tilde{t}}), \tilde{s}_{\tilde{t} + \tilde{\tau}}) \right]
            $$
            where $t_{0}$ and $\tilde{t}_{0}$ represent the current time of the SMDP on a goal level and task level, respectively.
            
            In other words, the objective is to find a function that maximizes the cumulative future reward on both levels by choosing an appropriate action $a$ in any given state $s$ at time $t$.
            Written in terms of the action-value function $Q$, the goal-level optimal policy for a given goal-level state $s$ and current time $t_{0}$ is defined as
            $$
                \pi_{\text{(goal)}}^{\star}(s_{t_{0}}) = \arg \max_{a \in \Omega_{t}} Q_{\text{(goal)}}^{\star}(s, a)
            $$
            Similarly, the task-level optimal policy for a goal $g$ at a given task-level state $\tilde{s}$ and current time $\tilde{t}_{0}$ is defined as
            $$
                \pi_{\text{(task)}}^{(g)\star}(\tilde{s}_{\tilde{t}_{0}}) = \arg \max_{\tilde{a} \in \Omega_{\tilde{t}}^{(g)}} Q_{\text{(task)}}^{(g)\star}(\tilde{s}, \tilde{a})
            $$
            
            Under the assumption that goal values are non-negative, the maximum potential value of cumulative rewards is obtained when \textit{all} goal and task deadlines in the to-do list are attained. We discuss this observation in details in Section~\ref{sec:inductive_biases}.

    \subsection{Maximizing the productivity of myopic workers by optimal gamification} \label{sec:optimalGamification}
    
        The SMDP defined above allows us to formalize a person's productivity from time $t_1$ to time $t_2$ by
        \begin{equation}
            \text{P}(s_{t_1},s_{t_2})= \frac{V^\star(s_{t_2})-V^\star(s_{t_1})}{t_2-t_1},
        \end{equation}
        where $s_{t_1}$ and $s_{t_2}$ describes the sets of tasks that the person had completed by time $t_1$ and time $t_2$ respectively.
        
        Following \cite{lieder2019cognitive}, we model people as \textit{myopic bounded agents} who generally choose tasks based on the difference between their immediate reward minus the subjective cost of working on a task, which we model as the product of the task's unpleasantness and its duration. That is, we assume that people select tasks according to the greedy policy
        \begin{equation}
            \pi_{\text{greedy}}(s) = \arg\underset{a}{\max} \, \mathds{E}\left[r(s,a,s')\right].
        \end{equation}
        
        Under this assumption, the problem of maximizing people's productivity by optimal gamification can be formalized as computing optimal incentives $f^\star(s,a)$ so that
        \begin{equation}
            \forall s: \pi^\star(s) \in \arg \underset{a}{\max} \left\{ f^\star(s,a)+\mathds{E}\left[r(s,a,s')\right] \right\}.    
        \end{equation}
        
        Lieder et al. \cite{lieder2019cognitive} proved that this can be achieved by setting $f(s,a)$ to the optimal incentives
        \begin{align} \label{eq:pseudoRewards}
            f^\star(s,a)
            & = \mathds{E}\left[ V^\star(s') | s, a \right] - V^\star(s) \\
            & = \max_{a'} Q^{\star}(s', a') - \max_{\tilde{a}} Q^{\star}(s, \tilde{a})
        \end{align}  

        To help people choose the most valuable task, their daily to-do list should include the tasks $\pi^\star(s_t), \pi^\star(s_{t+1}), \cdots $ that the optimal policy would choose on that particular day (where $t$ is the first time step of the person's work day).
        Unfortunately, naive computation of $f^\star$ is intractable for real-world applications. The goal of this text is to present a scalable algorithm for approximate computation of optimal incentives $f^\star$ for real-world applications of productivity apps.

\section{Solution}  \label{sec:solution}

\begin{figure}[!t]
        \centering
        \includegraphics[width=300px]{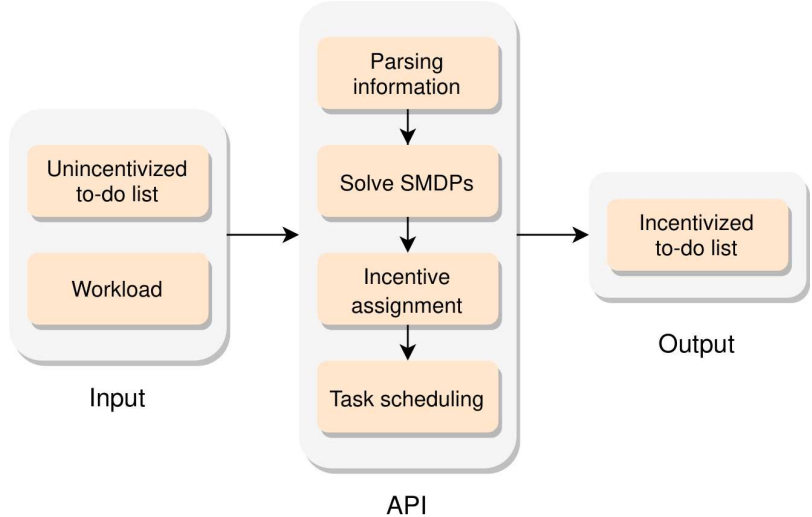}
        \caption{
            Complete pipeline.
            \textbf{Inputs:}
                \textit{Unincentivized to-do list} is a complete list of goals and tasks accompanied by all the relevant information such as goal values, time estimates, deadlines etc.;
                \textit{Workload} represents information on the desired amount of typical daily and today's working hours.
            \textbf{API:}
                \textit{Parsing information} is the procedure of parsing the provided to-do list information as text entries in a JSON dictionary;
                \textit{Solve SMDPs} is the procedure that solves the hierarchical structure of SMDPs;
                \textit{Incentive assignment} is the procedure of assigning incentives to tasks;
                \textit{Task scheduling} is the procedure that proposes a task list schedule.
            \textbf{Output:}
                \textit{Incentivized to-do list} is a gamified to-do list shown to the user.
        }
        \label{fig:pipeline}
    \end{figure}

    We start this section by introducing inductive biases (Section~\ref{sec:inductive_biases}) that lead to a drastic reduction in the problem complexity, tackling the problem of intractable computations for real-world to-do lists. Then, in the following sections, we describe the pipeline for computing an optimal gamified to-do list.
    That is, we describe the API components, their functionalities and 
    the procedure for generating a solution.
    In Section~\ref{sec:parsing}, we give a brief description of the procedure that parses user's input.
    Next, in Section~\ref{sec:algorithm}, we describe the procedure that constructs the hierarchy of goal- and task-level SMDPs and the relations between them, as well as the algorithmic procedure that solves them.
    Lastly, in Section~\ref{sec:incentivizingAndScheduling} we describe the procedure that assigns task incentives and the scheduling procedure that proposes a list of the most-valuable set of tasks to a user.
    An illustration of the complete procedure is given in Figure~\ref{fig:pipeline}.
    
    \subsection{Incorporating inductive biases}  \label{sec:inductive_biases}
    
        In order to further reduce the problem complexity, we introduce a set of assumptions that restrict the potential solution space by removing parts of it that are certainly sub-optimal. These kind of assumptions are formally known as \textit{inductive/learning biases} \cite{mitchell1980need}.
        We first describe an inductive bias in cases where to-do lists have no task deadlines. We then provide a proof that the inductive bias eliminates computations that cannot be part of an optimal solution. Finally, we extend the inductive bias to cases in which to-do lists contain task deadlines.
        As a consequence, the problem of to-do list gamification becomes solvable in a reasonable amount of time (see Section~\ref{sec:evaluations} for details).

        The inductive biases that we incorporate in the procedure of solving task-level SMDPs with constant non-zero loss rate $\lambda > 0$ (i.e. unit cost of working on a task) are:
        \begin{enumerate}
            \item If no task has a deadline (or not attaining one is not penalized), then any sequence of task execution is an optimal one.
            \item Tasks with deadlines should be executed according to the proximity of the deadlines in order to minimize the total penalty for not attaining those deadlines (if any).
        \end{enumerate}

        We support the claim for the first inductive bias via the following example. Let a goal $g \in \mathcal{G}$ have 3 tasks with time estimates $\{1, 2, 3\}$ such that \textit{none} of these actions correspond to the \textit{slack-off action}. Concretely, we show that the value of executing the sequence of actions $a_{1} \to a_{2} \to a_{3}$ is equivalent to the value of executing $a_{3} \to a_{2} \to a_{1}$ for arbitrary values of $\gamma$ and $\lambda$.
        
        \begin{align*}
            Q(s_{0}, a_{1})
            & = r(s_{0}, a_{1}, s_{1}) + \gamma^{1} Q(s_{1}, a_{2}, s_{3})  \\
            & = r(s_{0}, a_{1}, s_{1}) + \gamma^{1} \bigg( r(s_{1}, a_{2}, s_{3}) + \gamma^{2} Q(s_{3}, a_{3}, s_{6}) \bigg)  \\
            & = r(s_{0}, a_{1}, s_{1}) 
              + \gamma^{1} \bigg( r(s_{1}, a_{2}, s_{3}) 
              + \gamma^{2} \big(r(s_{3}, a_{3}, s_{6}) 
              + \gamma^{3} R(g) \cdot \Pi(\beta_{g}) \big) \bigg)  \\
            & = - \lambda \gamma^{0}
              + \gamma^{1} \bigg(
                - \lambda (\gamma^{0} + \gamma^{1}) + \gamma^{2} \big( 
                  - \lambda (\gamma^{0} + \gamma^{1} + \gamma^{2}) + \gamma^{3} R(g)  \cdot \Pi(\beta_{g})
                \big)
              \bigg)  \\
            & = - \lambda (\gamma^{0} + \gamma^{1} + \gamma^{2} + \gamma^{3} + \gamma^{4} + \gamma^{5}) + \gamma^{6} R(g) \cdot \Pi(\beta_{g}) \\
            & = - \lambda (\gamma^{0} + \gamma^{1} + \gamma^{2}) + \gamma^{3} \bigg(-\lambda (\gamma^{0} + \gamma^{1}) + \gamma^{2} \big( -\lambda \gamma^{0} + \gamma^{1} R(g) \cdot \Pi(\beta_{g}) \big) \bigg)  \\
            & = r(s_{0}, a_{3}, s_{3}) + \gamma^{3} \bigg( r(s_{3}, a_{2}, s_{5}) + \gamma^{2} \big( r(s_{5}, a_{1}, s_{6}) + \gamma^{1} R(g) \cdot \Pi(\beta_{g}) \big) \bigg)  \\
            & = r(s_{0}, a_{3}, s_{3}) + \gamma^{3} \bigg( r(s_{3}, a_{2}, s_{5}) + \gamma^{2} Q(s_{5}, a_{1}) \bigg)  \\
            & = r(s_{0}, a_{3}, s_{3}) + \gamma^{3} Q(s_{3}, a_{2}, s_{5})  \\
            & = Q(s_{0}, a_{3})  \\
            & = Q^{\star}(s_{0}, \cdot)
        \end{align*}
        where $s_{t}$ denotes the state $s$ at time $t$ and $a_{\tau}$ denotes an action $a$ with duration $\tau$. The equality above holds for all permuted sequences of action execution $a_{i_{1}} \to a_{i_{2}} \to a_{i_{3}}$ and generalization of this observation for higher number of actions and varying time estimates applies by induction.
        
        As a consequence of the first inductive bias, we are able to compute the Q-value for \textit{all} sequences of action execution by evaluating \textit{only one} (arbitrarily-chosen) sequence of actions. Given this inductive bias, the aim of solving task-level SMDPs becomes trivial since all sequences of actions are optimal ones.

        However, once task deadlines come into play, the solution loses its triviality. Executing an arbitrary task at each time step does \textit{not} guarantee attainability of task deadlines and thus optimality. Luckily, although this introduces an overhead expense in the procedure of computing an optimal task-level policy, dealing with it is not overly complicated. In this setting, where at least one task has a deadline, the objective is to minimize the penalty induced by not attaining a deadline in order to maximize the Q-value for a particular sequence of actions. Therefore, an optimal sequence of actions is one that executes tasks according to the temporal proximity of their deadlines. Consequently, the optimal policy at each step has to greedily choose the task with the next upcoming deadline (while breaking ties randomly, if any). Finally, the computation of optimal task-level policy requires two steps: (1) sort task-level actions in an increasing order according to their deadlines (for all goals separately); (2) greedily execute actions in that order. We present a complete algorithmic solution in Section~\ref{sec:algorithm}.
        
        Exact task incentivization requires computing Q-values for all uncompleted task-level actions at the present state. Doing this in a brute-force manner (i.e. without incorporating inductive biases) corresponds to exhaustive exploration of the state space and yields a computational time complexity of $O(2^{n})$, where $n$ is the total number of tasks on the to-do list. Incorporating inductive biases reduces the problem complexity from exponential -- $O(2^{n})$ -- to quadratic -- $O(n^{2})$. Additionally, sorting tasks according to task deadlines takes at most $O(n \log n)$ computational time, which is negligible compared to the other terms in the problem complexity.

     \subsection{Parsing information} \label{sec:parsing}
\begin{figure}[!t]
            \centering
            \includegraphics[width=\textwidth]{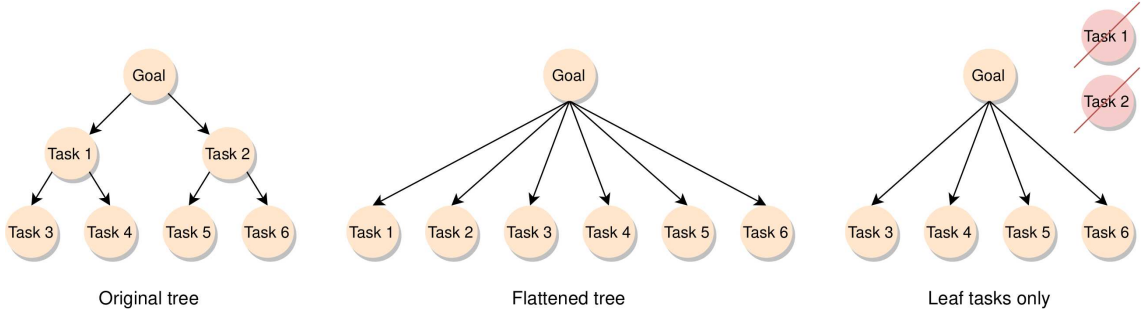}
            \caption{Different methods of parsing hierarchical structures.}
            \label{fig:parsing_hierarchy}
        \end{figure}

        The parsing component of the API converts an unincentivized to-do list (accompanied with additional information such as current tasks in that day's schedule) to information that can generate a hierarchy of SMDPs and solve them. The information goes through the following parsing sub-components before it reaches the algorithmic procedure:
        \begin{enumerate}
            \item \textbf{Parsing hierarchical structure.} Hierarchical to-do lists have goals on the main level and tasks on the sub-levels. Here, we provide two possible methods for parsing the hierarchical structure of a to-do list (see Figure~\ref{fig:parsing_hierarchy}).
            \begin{itemize}
                \item Flattening by taking into account the internal structure of the hierarchy.
                \item Discarding the internal structure by taking into account only the tasks that have no sub-tasks (i.e. leaf nodes in the to-do-list tree).
            \end{itemize}
            \item \textbf{Parsing goals and tasks.} This sub-component parses goal and task descriptions in order to extract information provided in them such as values, time estimates etc.
            \item \textbf{Parsing scheduling tags.} This sub-component parses information on the desired ``do" days/dates for each task on a to-do list. This information is necessary so that the API knows when to propose a given tasks, as well as how much working time is available for each day in the week.
            \item \textbf{Parsing deadlines.} Here, the API uses the previously-parsed information on today's workload and the typical day's workload in order to calculate the available working time (in minutes) until the specified deadline for each task that has one.
        \end{enumerate}
        
        Once the parsing procedure is done, the API checks whether all goal values and the average goal values (calculated as a fraction of the goal value over its total time estimate) are within predefined ranges. If both conditions are satisfied then the parsed information is forwarded to the procedure that generates and solves SMDPs. Otherwise, the user is asked to modify the inputted to-do list.

        \begin{figure}[!t]
            \centering
            \includegraphics[width=\textwidth]{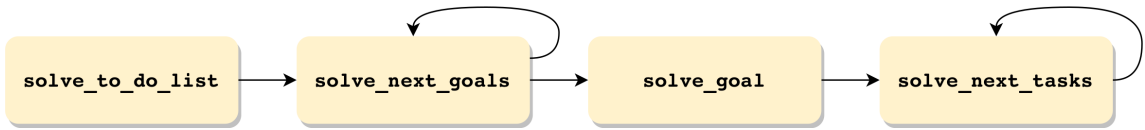}
            \caption{Algorithmic procedure.}
            \label{fig:procedure}
        \end{figure}
        
    \subsection{Solving SMDPs}  \label{sec:algorithm}
    
        In this section, we present an algorithmic procedure that efficiently solves a hierarchy of SMDPs. In general, the complete procedure is divided into four major methods which are executed in a hierarchical manner. Here, we give a brief description of their functionalities. A more detailed description of the procedure is provided in a form of a pseudo-code in Algorithms~\ref{alg:SolveToDoList}, \ref{alg:SolveNextGoals}, \ref{alg:SolveGoal}, and \ref{alg:SolveNextTasks}. A graphical representation of the complete procedure is provided in Figure~\ref{fig:procedure}.
        
        Initially, the \texttt{solve\_to\_do\_list} method (Algorithm~\ref{alg:SolveToDoList}) is called. This method initializes the global parameters (i.e. discount factor $\gamma$, loss rate $\lambda$, penalty rate $\psi$), incorporates the inductive biases defined in Section~\ref{sec:inductive_biases} by generating sorted lists of tasks and initiates the method for computing goal-level $Q$-values for each possible goal ordering (i.e. \texttt{solve\_next\_goals}). Once the goal-level $Q$-values are computed, the method computes the optimal policy of the goal-level SMDP $\pi_{\text{(goal)}}^{\star}$, collects all task-level actions, updates their Q-value for the initial task-level state $\tilde{s}_{t_{0}}$ with the Q-value of the succeeding optimal goal, and computes task incentives at initial time $t_{0}$.

        % ===== solve_to_do_list =====
        \begin{algorithm}[!p]
            \caption{Pseudo-code for the method that initiates computing goal-level $Q$-values, and computes the goal-level optimal policy and goal-level pseudo-rewards.}
            \label{alg:SolveToDoList}
            
            \TitleOfAlgo{\texttt{solve\_to\_do\_list}($s$, $t_{0}$, $\mathcal{G}$, $\gamma$, $\lambda$, $\psi$)}
            
            \SetKwInOut{Input}{Input}

            \Input{Initial state $s$ \\
                   Initial time $t_{0}$ \\
                   Set of goals $\mathcal{G}$ \\
                   Discount factor $\gamma$ \\
                   Loss rate $\lambda$ \\
                   Penalty rate $\psi$
            }

            \DontPrintSemicolon
            \;
            
            Set $\gamma$, $\lambda$, $\psi$ as global variables  \;
            
            \ForEach{ goal $g \in \mathcal{G}$ }{
                Sort tasks by deadlines and store them in $q^{(g)}$  \;
            }
            
            \;
            
            \texttt{solve\_next\_goals}($s$, $t_{0}$, $\mathcal{G}$) (Solve to-do list recursively)  \;
            
            $\pi_{\text{(goal)}}^{\star}(s_{t_{0}}) \leftarrow \arg \underset{ a \in \Omega_{t_{0}} }{ \max } \ Q_{\text{(goal)}}^{\star}(s_{t_{0}}, a)$ (Store optimal policy)  \;
            
            \;
            
            $\mathcal{A} \leftarrow \{ \}$ (Initialize collection of task-level actions)  \;
            \ForEach{ goal $g \in \mathcal{G}$ }{
            
                Combine task $Q$-values with the $Q$-value of the next optimal goal \;
                
                \ForEach{ $\tilde{a} \in \mathcal{A}$ }{
                    $Q^{(g)\star}_{\text{(task)}}(\tilde{s}_{t_{0}}, \tilde{a})
                    \leftarrow Q_{\text{(task)}}^{(g)\star}(\tilde{s}_{t_{0}}, \tilde{a}) + \underset{a' \in \Omega_{t'}}{ \max } \, Q_{\text{(goal)}}^{\star}(s'_{t'}, a')$
                }

                $\mathcal{A} \leftarrow \mathcal{A} \cup \mathcal{A}_{\text{(task)}}^{(g)}$ (Update collection of tasks)
            }
            \;
            
            Compute incentives $r'(\cdot, \cdot, \cdot)$ for all tasks $\mathcal{A}$ at $t_{0}$  \;
            
        \end{algorithm}

        The \texttt{solve\_next\_goals} method (Algorithm~\ref{alg:SolveNextGoals}) is initiated by the \texttt{solve\_to\_do\_list} method and provided with the information of the current goal-level state $s$, current goal-level time $t$ and the set of uncompleted goals $\mathcal{UG}$ (which is initially the complete set of goals $\mathcal{G}$). This method computes the $Q$-values for all possible goal orderings (i.e. goal-level state-time pairs $s_{t}$) by solving each individual goal via the \texttt{solve\_goal} method in a recursive manner. Since the method computes $Q$-values for all possible permutations of goal orderings, its computational time is $O(| \mathbf{\mathcal{G}} | !)$, where $| \mathbf{\mathcal{G}} |$ is the total number of goals in a to-do list.
        
        % ===== get_next_goal =====
        \begin{algorithm}[!p]
            \caption{Pseudo-code for the method that computes goal-level $Q$-values.}
            \label{alg:SolveNextGoals}
            
            \TitleOfAlgo{\texttt{solve\_next\_goals}($s$, $t$, $\mathcal{UG}$)}
            
            \SetKwInOut{Input}{Input}

            \Input{Current state $s$;  \\
                   Current time $t$;  \\
                   Set of uncompleted goals $\mathcal{UG}$;  \\
            }

            \DontPrintSemicolon
            \;
            
            \eIf{ $\mathcal{UG} \neq \emptyset $ }{

                \ForEach{ uncompleted goal g $\in \mathcal{UG}$ }{
                
                    $\tau \leftarrow$ (Get goal time estimate) \;
                    
                    $t' \leftarrow t + \tau$  (Move $\tau$ time steps in the future)  \;
                    
                    $s_{t'}' \leftarrow$ (Move to next state by taking action $a$ in state $s_{t}$)  \;
                    
                    \;
                    
                    \If{ $Q_{\text{(goal)}}^{\star}(s_{t}, g)$ not computed }{
                        
                        \texttt{solve\_goal}($g$, $t$)  \;
                        
                        \texttt{solve\_next\_goals}($s'$, $t'$, $\mathcal{UG} - \{ \texttt{g} \}$) (Make a recursive call )  \;
                    }
                    
                    \;
                    
                    $Q_{\text{(goal)}}^{\star}(s_{t}, g)
                    \leftarrow r(s_{t}, g, s_{t'}')
                    + \gamma^{\tau} \underset{a' \in \Omega_{t'}}{ \max } \, Q_{\text{(goal)}}^{\star}(s_{t'}', a') $ (Update $Q$-value) \;
                    
                    $\pi_{\text{(goal)}}^{\star}(s_{t'}') \leftarrow \arg \underset{ a' \in \Omega_{t'} }{ \max } \ Q_{\text{(goal)}}^{\star}(s_{t'}', a')$ (Store optimal policy)  \;
                }
            }
            {
                $Q_{\text{(goal)}}^{\star}(s_{t}, \perp) \leftarrow 0$ (Initialize terminal state)
            }
            
        \end{algorithm}

        The \texttt{solve\_goal} method (Algorithm~\ref{alg:SolveGoal}) initiates the recursive procedure of computing task-level $Q$-values for a given goal ${g}$ at a given initial time $\tilde{t}_{0}$. Its purpose is to to compute expected future rewards for all task-level actions that might appear on the user's schedule for today (i.e. at present time $\tilde{t}_{0} = 0$) given the inductive biases defined in Section~\ref{sec:inductive_biases}. The computation of the task-level $Q$-values is done completely within the \texttt{solve\_next\_tasks} method. Once the task-level $Q$-values for the given goal are computed, the method computes the optimal task-level policy $\pi_{\text{(task)}}^{\star (g)}$. The computational cost of this method is $O(\tilde{n})$, where $\tilde{n}$ is the total number of tasks within the goal.
        
        % ===== solve_goal =====
        \begin{algorithm}[!t]
            \caption{Pseudo-code for the method that initiates computing task-level $Q$-values, and computes task-level optimal policy and task-level pseudo-rewards for a given goal.}
            \label{alg:SolveGoal}
            
            \TitleOfAlgo{\texttt{solve\_goal}($g$, $\tilde{t}_{0}$)}
            
            \SetKwInOut{Input}{Input}

            \Input{Goal $g$;  \\
                    Initial time $\tilde{t}_{0}$;  \\
            }

            \DontPrintSemicolon
            \;
            
            $\tilde{s} \leftarrow$ (Initialize task-level state)  \;
            
            $q^{(g)} \leftarrow$ (Get list of sorted tasks by deadlines) \;
            \, \;

            \eIf{ $t = 0$ }{
                \ForEach{ $\tilde{a} \in q$ }{
                    \texttt{solve\_next\_tasks}($\tilde{s}$, $\tilde{t}_{0}$, $q^{(g)}$, $0$, $\tilde{a}$)
                }
            }{
                \texttt{solve\_next\_tasks}($\tilde{s}$, $\tilde{t}_{0}$, $q^{(g)}$, $0$, /)
            }
            \, \;
            
            $\pi_{\text{(task)}}^{\star (g)}(\tilde{s}_{\tilde{t}_{0}}) \leftarrow \arg \underset{ \tilde{a} \in \Omega_{\tilde{t}_{0}}^{(g)} }{ \max } \ Q_{\text{(task)}}^{\star (g)}(\tilde{s}_{\tilde{t}_{0}}, \tilde{a})$
            \;
            
        \end{algorithm}
        
        The \texttt{solve\_next\_tasks} method (Algorithm~\ref{alg:SolveNextTasks}) is  initiated by the \texttt{solve\_goal} method and provided with the information of the current task-level state $\tilde{s}$, current task-level time $\tilde{t}$, sorted list of tasks $q$, index that keeps track of the solved tasks in the list $i$, as well as an optional task-level action $\tilde{a}$ to be executed next. This method computes the $Q$-value for a path in the restricted (by the inductive biases) set of possible paths. It starts by searching for the next uncompleted action in the sorted list of actions. If such a task exists, it makes the necessary state and time transitions and it makes a recursive call from the next state and time. If no such task exists, this absence implies that a terminal state has been reached, in which case the procedure stores the (potentially penalized) goal reward as a $Q$-value for that state and time. The computational cost of this method is $O(\tilde{n}^{b})$, where $\tilde{n}$ is the total number of tasks within the goal and $b$ is a branching factor encoding the number of possible time transitions.
        
        % ===== get_next_task =====
        \begin{algorithm}[!t]
            \caption{Pseudo-code for the method for computing task-level $Q$-values recursively.}
            \label{alg:SolveNextTasks}
        
            \TitleOfAlgo{\texttt{get\_next\_task}($\tilde{s}$, $\tilde{t}$, $q$, $i$, $\tilde{a}$)}
            
            \SetKwInOut{Input}{Input}

            \Input{Current state $\tilde{s}$;  \\
                    Current time $\tilde{t}$;  \\
                    List of sorted tasks by deadline $q$;  \\
                    Index for the list of sorted tasks by deadlines $i$;  \\
                    Next action $\tilde{a}$ (optional);  \\
            }

            \DontPrintSemicolon
            \;
            
            \If{ $\tilde{a}$ is not provided }{
                \While{ $i < \text{length}(q)$ }{
                    $\tilde{a} \leftarrow$ (Get the next ($i$-th) task from $q$)  \;
                    
                    \eIf{$\tilde{a}$ is completed}{
                        $i \leftarrow i + 1$  \;
                    }{
                        break  \;
                    }
                }
            }
            \;

            \eIf{ $\exists$ uncompleted task $\tilde{a}$ }{
            
                $d \leftarrow$ get task time deadline  \;
                
                $\mathcal{T} \leftarrow$ (Get all potential time transitions and corresponding probabilities)  \;
                
                $Q_{\text{(task)}}^{(g)}(\tilde{s}_{\tilde{t}}, \tilde{a}) \leftarrow 0$ (Initialize $Q$-value)  \;
                
                \;
                
                \ForEach{ (Transition time $\tilde{\tau}$, Transition probability $\tilde{p}$) $\in \mathcal{T}$}{
                
                    $\tilde{t}' \leftarrow \tilde{t} + \tilde{\tau}$ 
                    (Move $\tilde{\tau}$ time steps in the future)  \;
                    
                    $\tilde{s}_{\tilde{t}'}' \leftarrow$
                    (Move to next state by taking action $\tilde{a}$ in state $\tilde{s}_{\tilde{t}}$)  \;
                
                    $r(\tilde{s}_{\tilde{t}}, \tilde{a}, \tilde{s}_{\tilde{t}'}') \leftarrow - \sum_{k=0}^{\tilde{\tau}-1} \gamma^{k} \lambda$
                    (Calculate total loss for action $\tilde{a}$)  \;
                    
                    $\beta_{g} \leftarrow \beta_{g} + \mathbb{I}(\tilde{t}' > d) \cdot (\tilde{t}' - d) \cdot \psi$
                    (Update penalty if missed deadline)  \;
                    
                    \;
                    
                    \texttt{get\_next\_task}($\tilde{s}'$, $\tilde{t}'$, $q$, $i$, /) \;
                    
                    \;

                    $
                        Q_{\text{(task)}}^{(g)}(\tilde{s}_{\tilde{t}}, \tilde{a}) 
                        \leftarrow Q_{\text{(task)}}^{(g)}(\tilde{s}_{\tilde{t}}, \tilde{a}) + \tilde{p} \, \cdot
                        \left[ 
                            r(\tilde{s}_{\tilde{t}}, \tilde{a}, \tilde{s}_{\tilde{t}'}') 
                            + \gamma^{\tilde{\tau}} \cdot \underset{ \tilde{a}' }{ \max } \ Q_{\text{(task)}}^{(g)}( \tilde{s}_{\tilde{t}'}', \tilde{a}' )
                        \right]
                    $  \;
                        
                    $\pi_{\text{(task)}}^{(g)}(\tilde{s'}_{\tilde{t}'}) \leftarrow \arg \underset{ \tilde{a}'
                    }{ \max } \ Q_{\text{(task)}}^{(g)}(\tilde{s'}_{\tilde{t}'}, \tilde{a}')$
                }
            }
            {
                $Q_{\text{(task)}}^{(g)}(\tilde{s}_{\tilde{t}}, \tilde{\perp}) \leftarrow R(g) \cdot \Pi(\beta_{g})$ (Initialize terminal state)  \;
            }
            
        \end{algorithm}

    \subsection{Incentivizing and scheduling tasks}  \label{sec:incentivizingAndScheduling}
    
        The aim of this work is to develop a gamified to-do list app that helps people overcome the motivational obstacles that result from the misalignment between the immediate cost of work and its long-term benefits.  For that matter, we leverage the method of optimal gamification presented in Section~\ref{sec:optimalGamification} to compute pseudo-rewards $f(s_{t}, a, s_{t'}')$ for completing task $a$ in state $s$ at time $t$ leading to state $s'$ at time $t'$. We compute those point values by combining the pseudo-rewards computed by the optimal reward shaping method by Ng et al. \cite{ng1999policy} with the original reward $r(s_{t}, a, s_{t'}')$ for performing the task, that is
        $$
            f(s_{t}, a) = f'(s,a) + \mathbb{E} \left[ r(s_{t}, a, s_{t'}') \right]
        $$
        where $f'(s,a) = m \cdot f^{\star}(s_{t}, a)+b$ is an optimality-preserving linear function of the optimal pseudo-rewards $f^\star(s,a)$ defined in Equation~\ref{eq:pseudoRewards}. Adding the rewards $r$ to the pseudo-rewards is also optimality preserving because this is equivalent to multiplying the reward function by $2$ before adding the optimality preserving pseudo-rewards. Concretely, we use the following linear transformation in order to ensure that the sum of all pseudo-rewards equals the sum of goal values. To break ties between multiple optimal tasks by preferring longer tasks over shorter ones, we set the scaling parameter $m = 1.1$ and we derive the value for the bias parameter $b$ in the following manner:

        \begin{align*}
            \sum_{a} \left( m \cdot f^{\star}(s_{0}, a) + b + \mathbb{E} \left[ r(s_{0}, a, s'_{t'}) \right] \right) & = \sum_{g \in \mathcal{G}} R(g)  \\
              m \cdot \sum_{a} f^{\star}(s_{0}, a) + \sum_{a} b + \sum_{a} \mathbb{E} \left[ r(s_{0}, a, s'_{t'}) \right] & = \sum_{g \in \mathcal{G}} R(g)  \\
              m \cdot \sum_{a} f^{\star}(s_{0}, a) + n \cdot b + \sum_{a} \mathbb{E} \left[ r(s_{0}, a, s'_{t'}) \right] & = \sum_{g \in \mathcal{G}} R(g)
        \end{align*}
        $$
            b = \dfrac{\sum_{g \in \mathcal{G}} R(g) - m \cdot \sum_{a} f^{\star}(s_{0}, a) - \sum_{a} \mathbb{E} \left[ r(s_{0}, a, s'_{t'}) \right] }{ n }
        $$

        \vspace{0.3cm}
        
        Once the pseudo-rewards are computed, the scheduling procedure acquires the most-valuable set of tasks that do not exceed the desired user workload as well as other tasks to be scheduled for the present day, rounds their incentives to a pre-defined number of decimals, and proposes them as next tasks to be executed.

\section{API}  \label{sec:api}

    The API that we have developed can serve as a back-end to any to-do list gamification application. It is available online at \href{https://github.com/RationalityEnhancementGroup/todolistAPI}{\textbf{https://github.com/RationalityEnhancementGroup/todolistAPI}}. Details on how to communicate with the API can be found in the \texttt{README.md} file there, and we briefly describe the communication in the following paragraphs and the Appendix.
    
    In our specific case, we coupled the API with a research version of the productivity application named CompliceX\footnote{\textbf{https://reg.complicex.science/}}, which has been developed especially for to-do-list gamification. The communication between the gamification application and the API occurs in the following manner:
    \begin{enumerate}
        \item The gamification application sends a POST request with a specific URL to the API hosted on Heroku\footnote{\textbf{https://safe-retreat-20317.herokuapp.com/}},
        which provides information about the to-do list in JSON format. Details on the format expected as input are provided in Section~\ref{sec:api_input}.
        \item After the POST request is received, the API parses the provided to-do-list information, computes task incentives, proposes a daily schedule of incentivized tasks, and sends this information back to the gamification application in JSON format. Details on the output that the API returns are provided in Section~\ref{sec:api_output}.
    \end{enumerate}
    
    Additionally, CompliceX communicates with a productivity application named WorkFlowy\footnote{We recommend \texttt{opusfluxus} as a NodeJS wrapper for WorkFlowy. The source code is available on GitHub - \textbf{https://github.com/malcolmocean/opusfluxus}}, but we omit describing their communication since the API communicates exclusively with CompliceX. Examples of input and output in JSON format can be found in the \inlinecode{examples/use\_cases} folder of the project's GitHub repository.

\section{Evaluations}  \label{sec:evaluations}
    
    \subsection{Inspecting pseudo-rewards}
        In order to ensure that the computed task incentives are meaningful in real-world applications, we assess them according to the following criteria:
        \begin{itemize}
            \item An optimal task is assigned the maximum number of points.
            \item Incentives of optimal tasks at current time are as high as or higher than what they were in the previous time step.
            \item The number of points and their differences across tasks motivate the selection of one task over others.
        \end{itemize}
        
        To illustrate this, we provide a use case of a student's to-do list with 3 long-term goals. Figure~\ref{fig:use_case_input} shows an unincentivized to-do list with all details about its goals and tasks. Figure~\ref{fig:use_case_output} shows an incentivized to-do list, in which the first 4 tasks are scheduled for the current day according to the desired workload, as well as updated incentive values after the next optimal task has been completed.
        For this use case example, our method used 2 possible task durations and scaled the user-provided time estimates by a planning fallacy constant of $1.39$.
        As a result, we show the scaled time estimates to the user in order to make the user aware of potential discrepancies between the estimated time and the real-world task-execution time.
        Moreover, the incentives are computed by optimal gamification and shifted such that they convey the long-term value of performing the task (as defined in Section~\ref{sec:incentivizingAndScheduling}).
        The incentives are scaled to the closest integer and their broad range ensures that the incentives are scaled up with the mental effort needed to execute these tasks.

    \subsection{Speed and reliability tests in different scenarios}  \label{sec:evalAPI}
    
        To show that the API is scalable in various real-world scenarios, we deployed the API on \href{https://www.heroku.com/}{Heroku} and tested its speed and reliability as performance measures. We measure reliability by the proportion of trials in which the API responds without throwing a \textit{timeout} error. The default timeout for a request on Heroku is 30 seconds. Since we want users to be able to interact with our API in real time, we set 28 seconds as an upper limit for the API to process a request.
        Concretely, we tested the API for for various number of daily working hours (i.e. 8, 12, 16), number of goals (i.e. 1 to 10), number of tasks per goal (i.e. 10 to 250), and number of possible task durations per task (i.e. 1 and 2). For simplicity, we fixed the average task time duration to be 15 minutes, which corresponds to 32, 48 and 64 daily tasks for 8, 12, and 16 daily working hours, respectively.
        
\begin{figure}[!hp]
            \centering
            \includegraphics[width=0.8\textwidth]{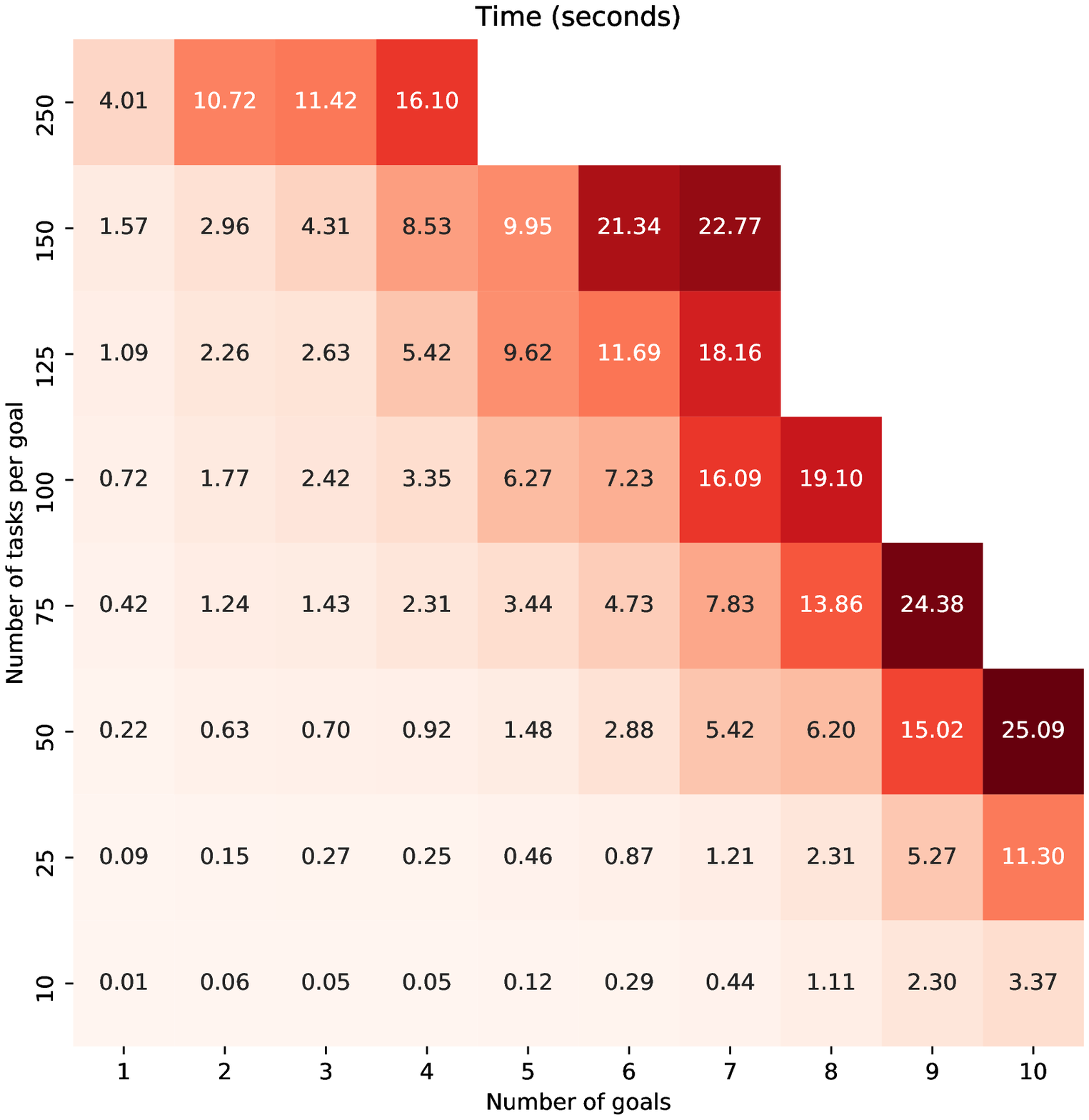}
            \caption{Speed tests for scenario: 8 working hours, 15 minutes average task time estimate, 1 possible task duration.}
            \label{fig:1b_480tm_15te_time}
        \end{figure}

\begin{figure}[!hp]
            \centering
            \includegraphics[width=0.8\textwidth]{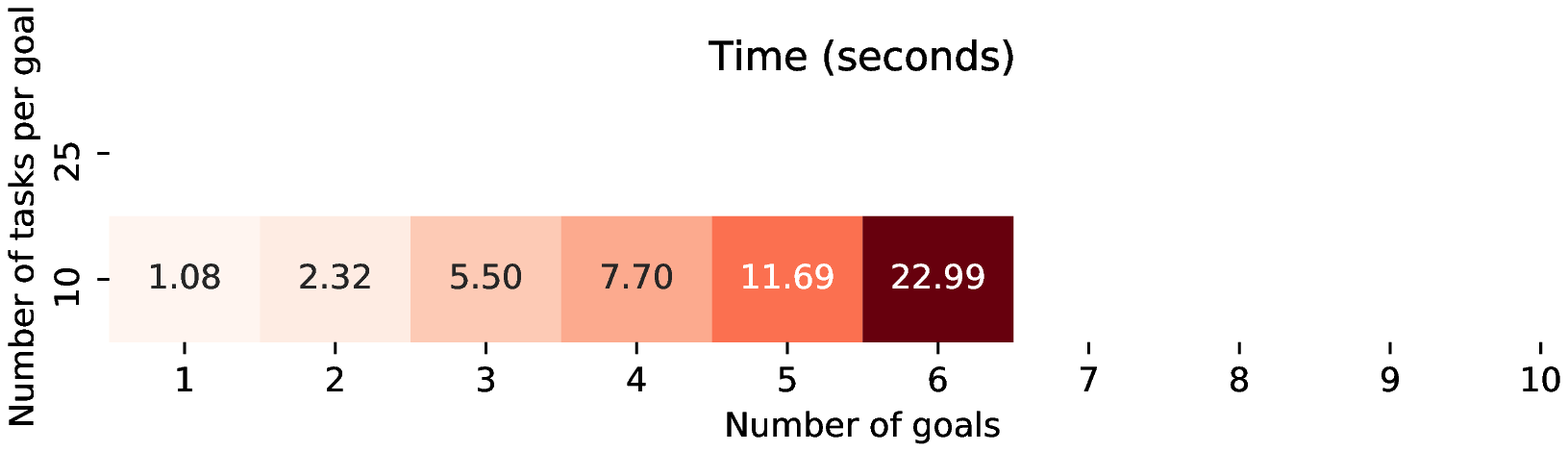}
            \caption{Speed tests for scenario: 8 working hours, 15 minutes average task time estimate, 2 possible task durations.}
            \label{fig:2b_480tm_15te_time}
        \end{figure}

\begin{figure}[!hp]
            \centering
            \includegraphics[width=0.8\textwidth]{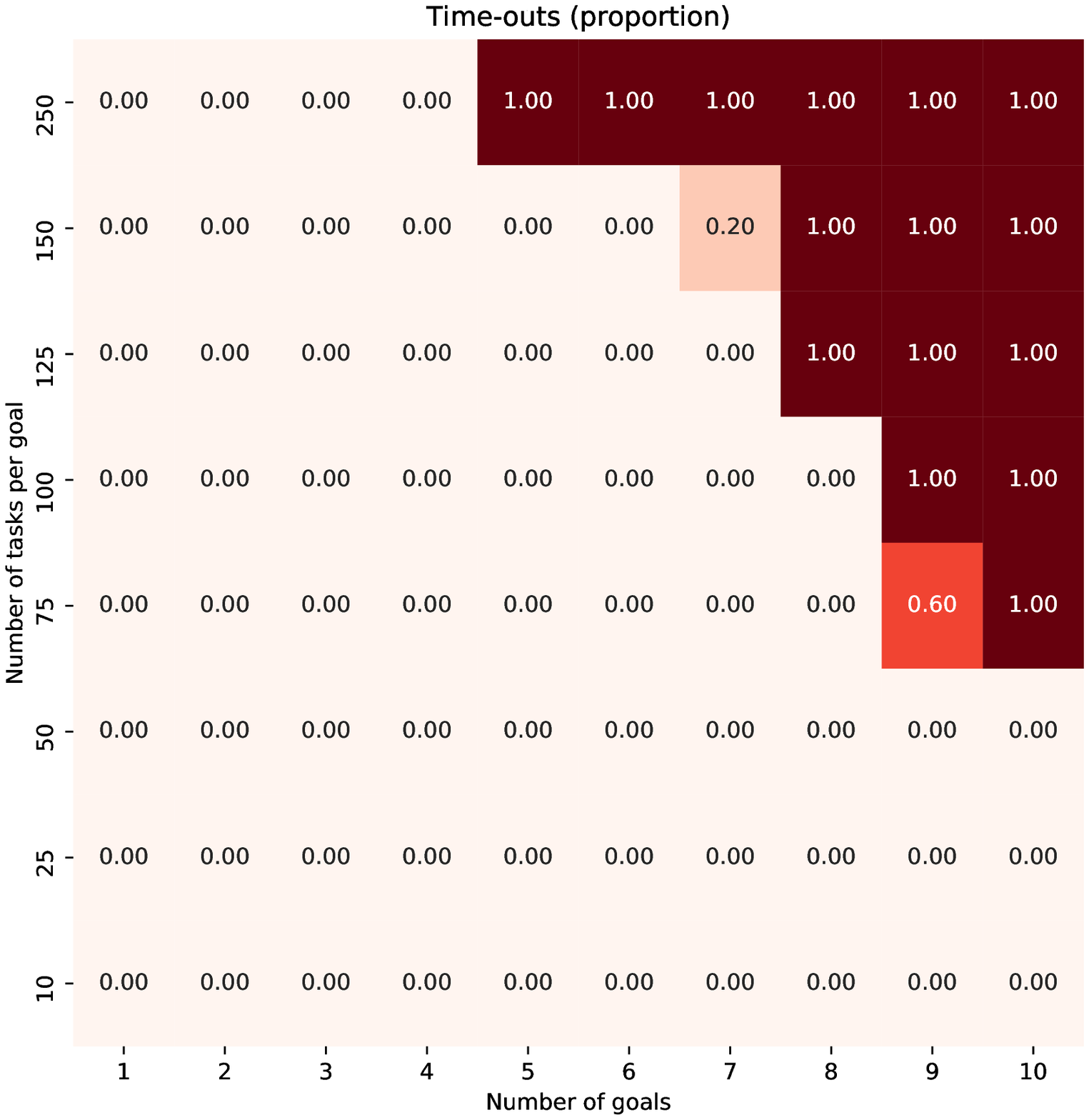}
            \caption{Reliability tests for scenario: 8 working hours, 15 minutes average task time estimate, 1 possible task duration.}
            \label{fig:1b_480tm_15te_tout}
        \end{figure}

\begin{figure}[!hp]
            \centering
            \includegraphics[width=0.8\textwidth]{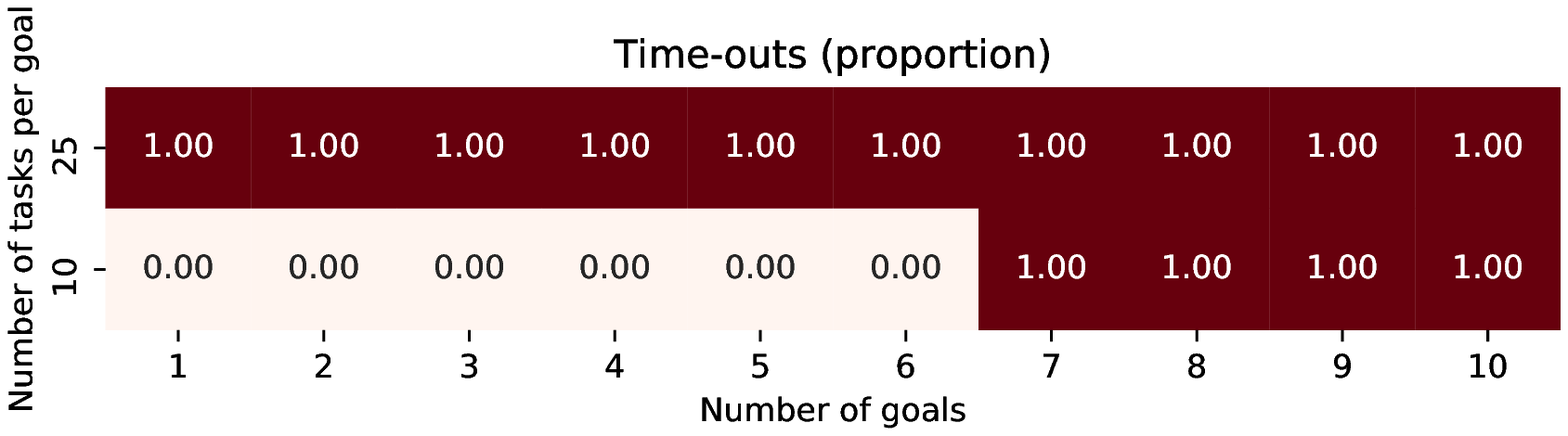}
            \caption{Reliability tests for scenario: 8 working hours, 15 minutes average task time estimate, 2 possible task durations.}
            \label{fig:2b_480tm_15te_tout}
        \end{figure}

\begin{figure}[!hp]
            \centering
            \includegraphics[width=0.8\textwidth]{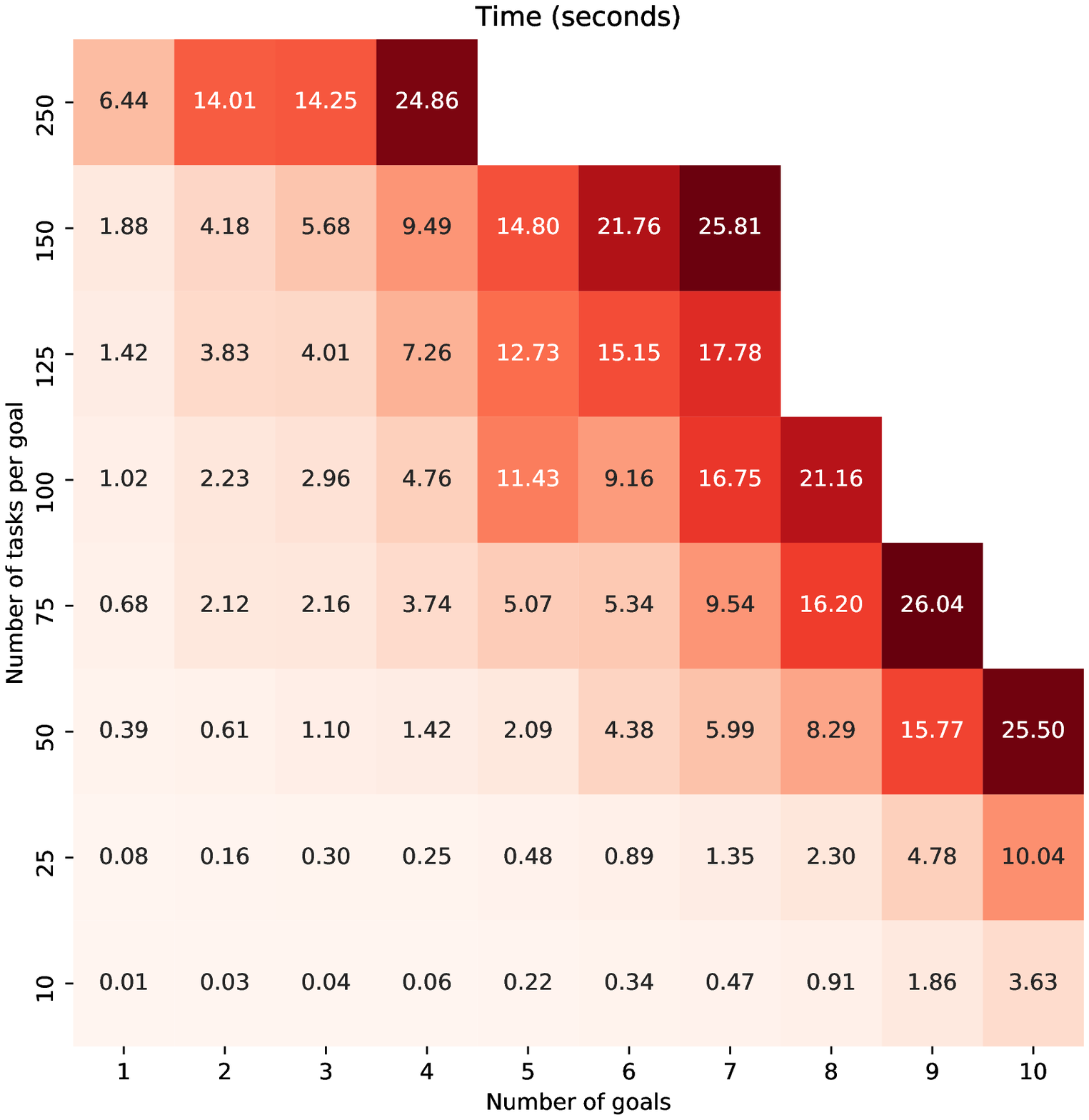}
            \caption{Speed tests for scenario: 12 working hours, 15 minutes average task time estimate, 1 possible task duration.}
            \label{fig:1b_720tm_15te_time}
        \end{figure}

\begin{figure}[!hp]
            \centering
            \includegraphics[width=0.8\textwidth]{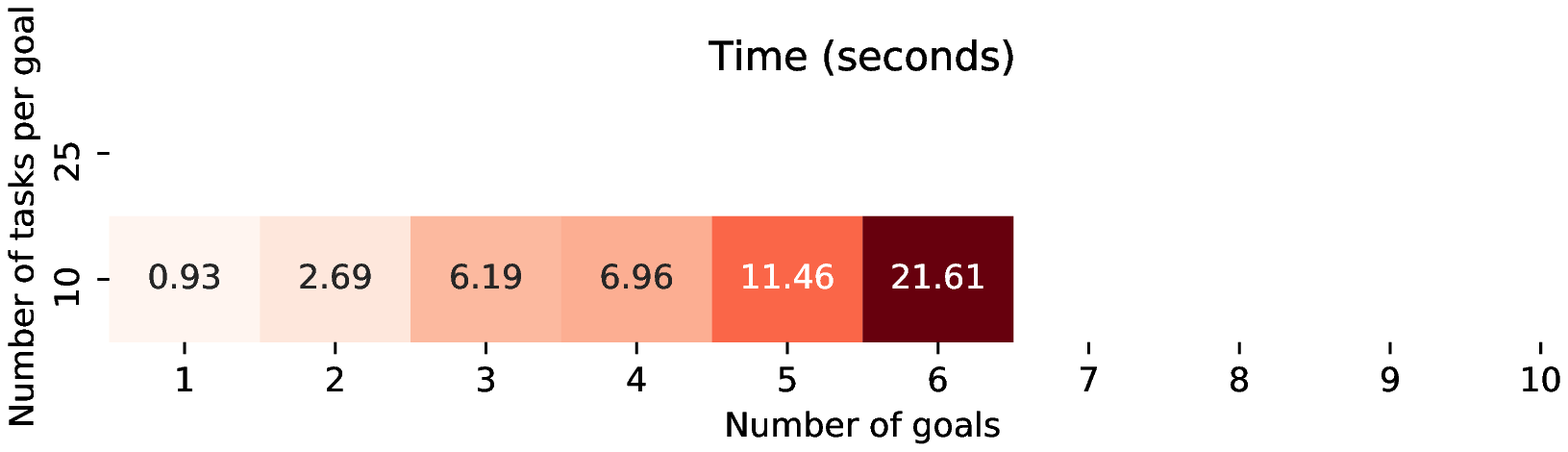}
            \caption{Speed tests for scenario: 12 working hours, 15 minutes average task time estimate, 2 possible task durations.}
            \label{fig:2b_720tm_15te_time}
        \end{figure}

\begin{figure}[!hp]
            \centering
            \includegraphics[width=0.8\textwidth]{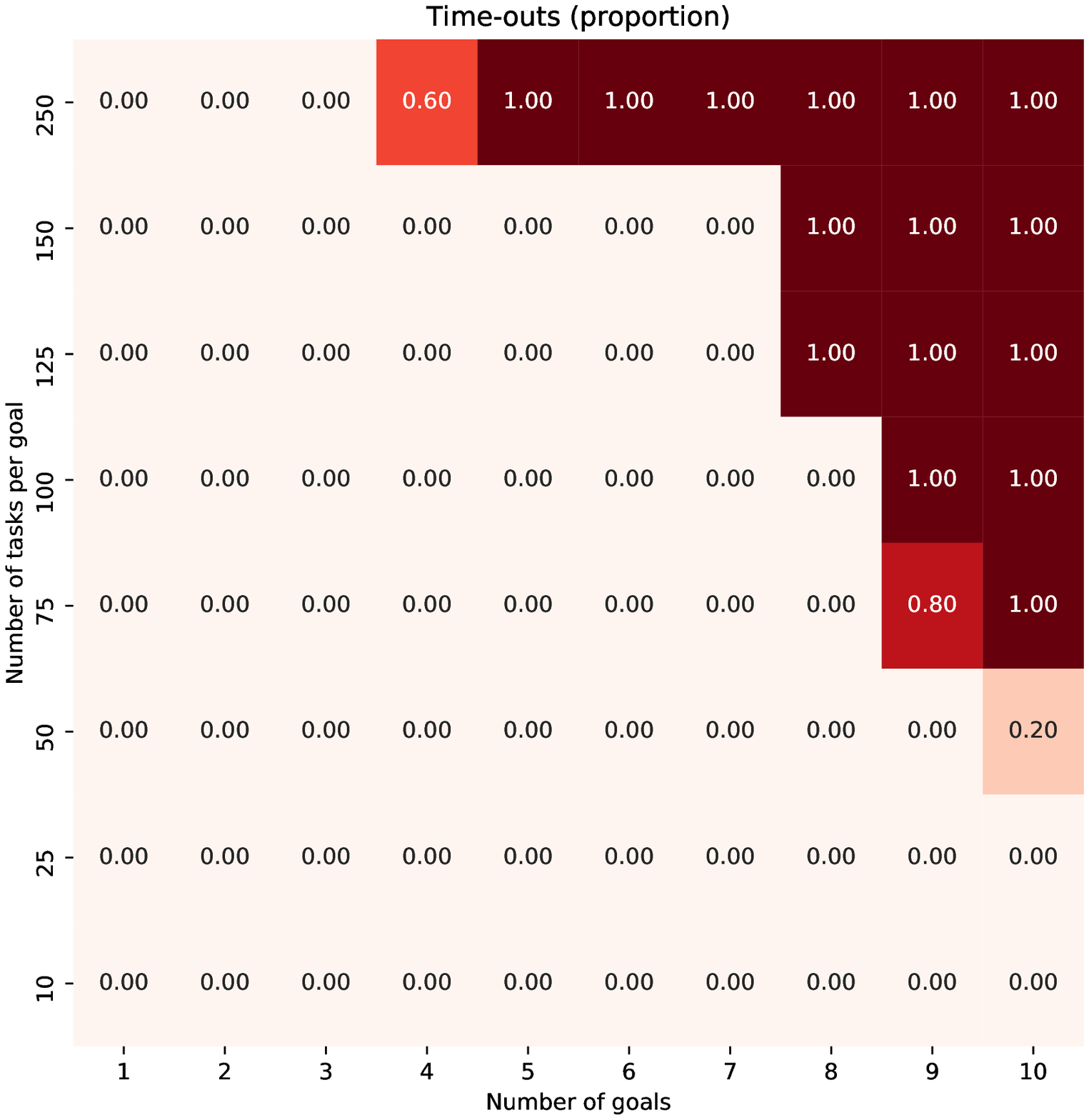}
            \caption{Reliability tests for scenario: 12 working hours, 15 minutes average task time estimate, 1 possible task duration.}
            \label{fig:1b_720tm_15te_tout}
        \end{figure}

\begin{figure}[!hp]
            \centering
            \includegraphics[width=0.8\textwidth]{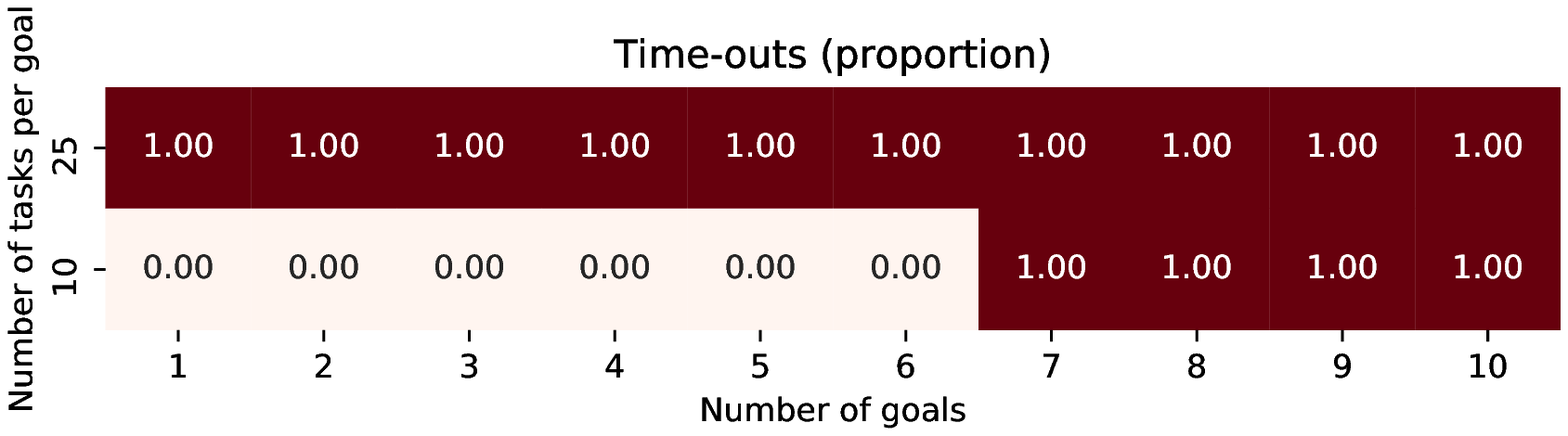}
            \caption{Reliability tests for scenario: 12 working hours, 15 minutes average task time estimate, 2 possible task durations.}
            \label{fig:2b_720tm_15te_tout}
        \end{figure}

\begin{figure}[!hp]
            \centering
            \includegraphics[width=0.8\textwidth]{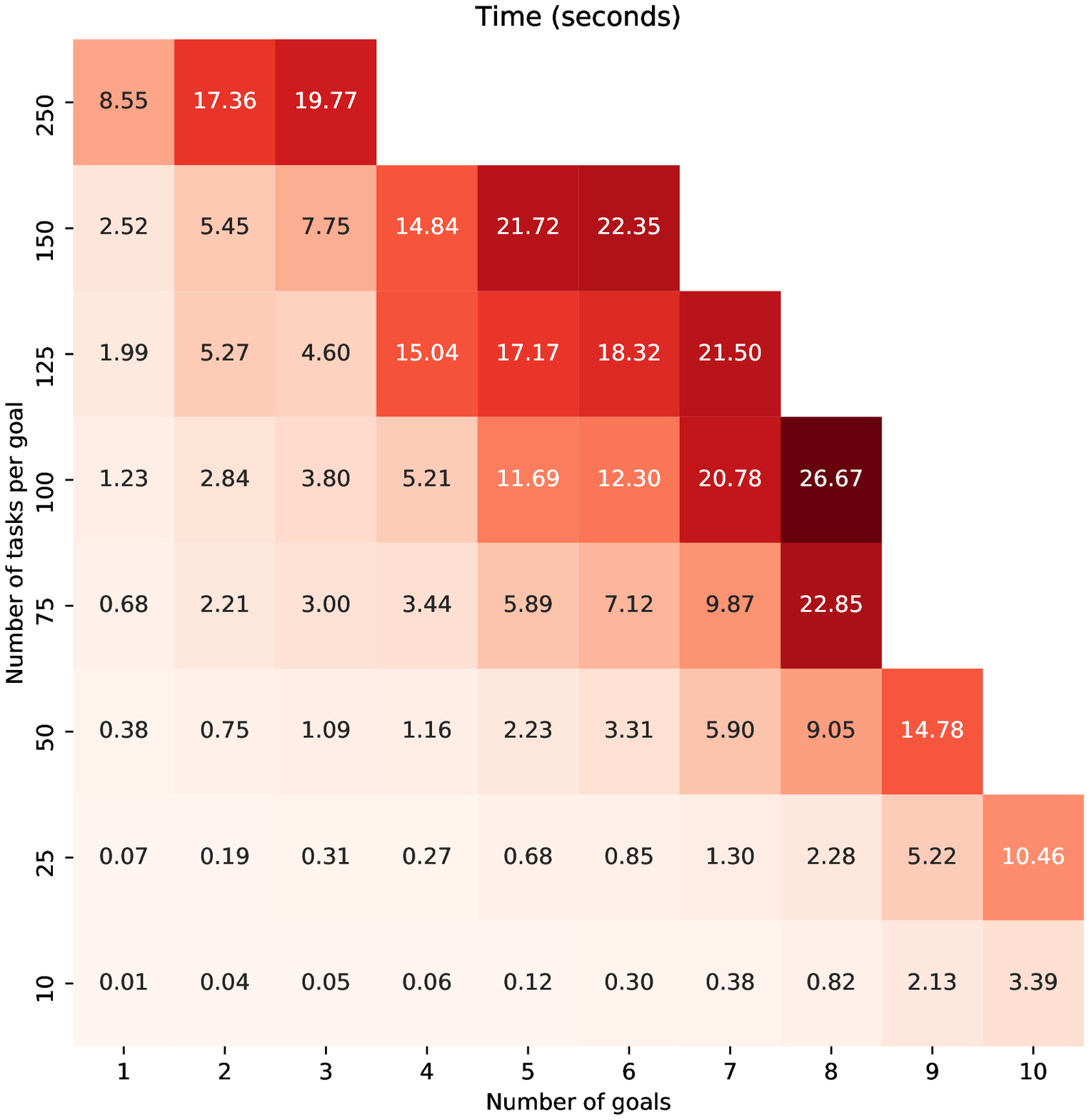}
            \caption{Speed tests for scenario: 16 working hours, 15 minutes average task time estimate, 1 possible task duration.}
            \label{fig:1b_960tm_15te_time}
        \end{figure}

\begin{figure}[!hp]
            \centering
            \includegraphics[width=0.8\textwidth]{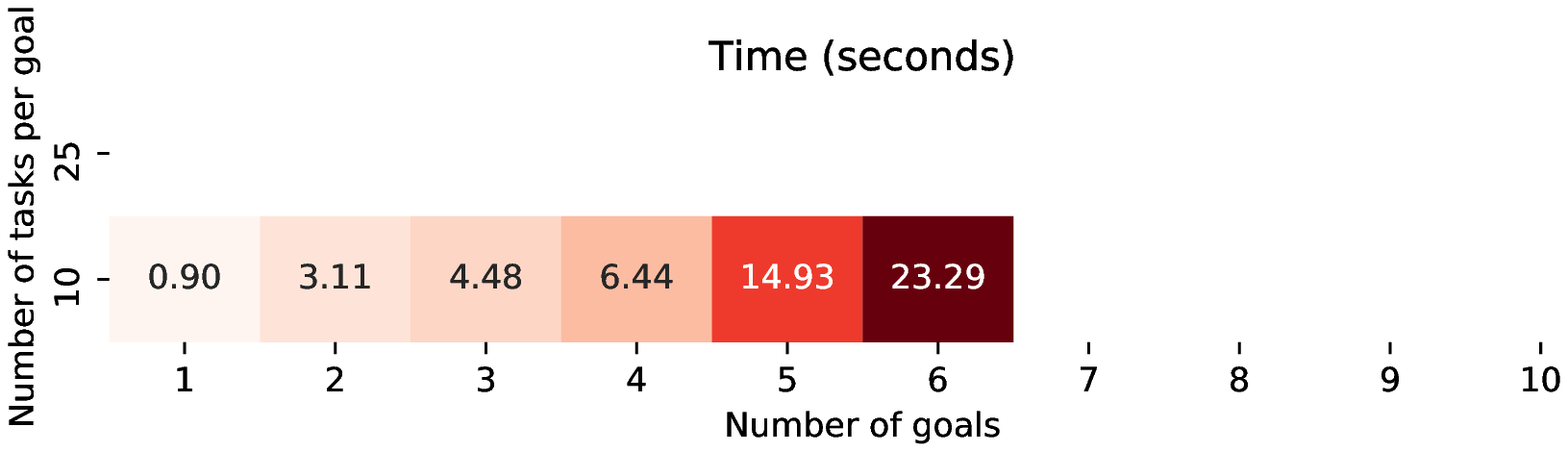}
            \caption{Speed tests for scenario: 16 working hours, 15 minutes average task time estimate, 2 possible task durations.}
            \label{fig:2b_960tm_15te_time}
        \end{figure}

\begin{figure}[!hp]
            \centering
            \includegraphics[width=0.8\textwidth]{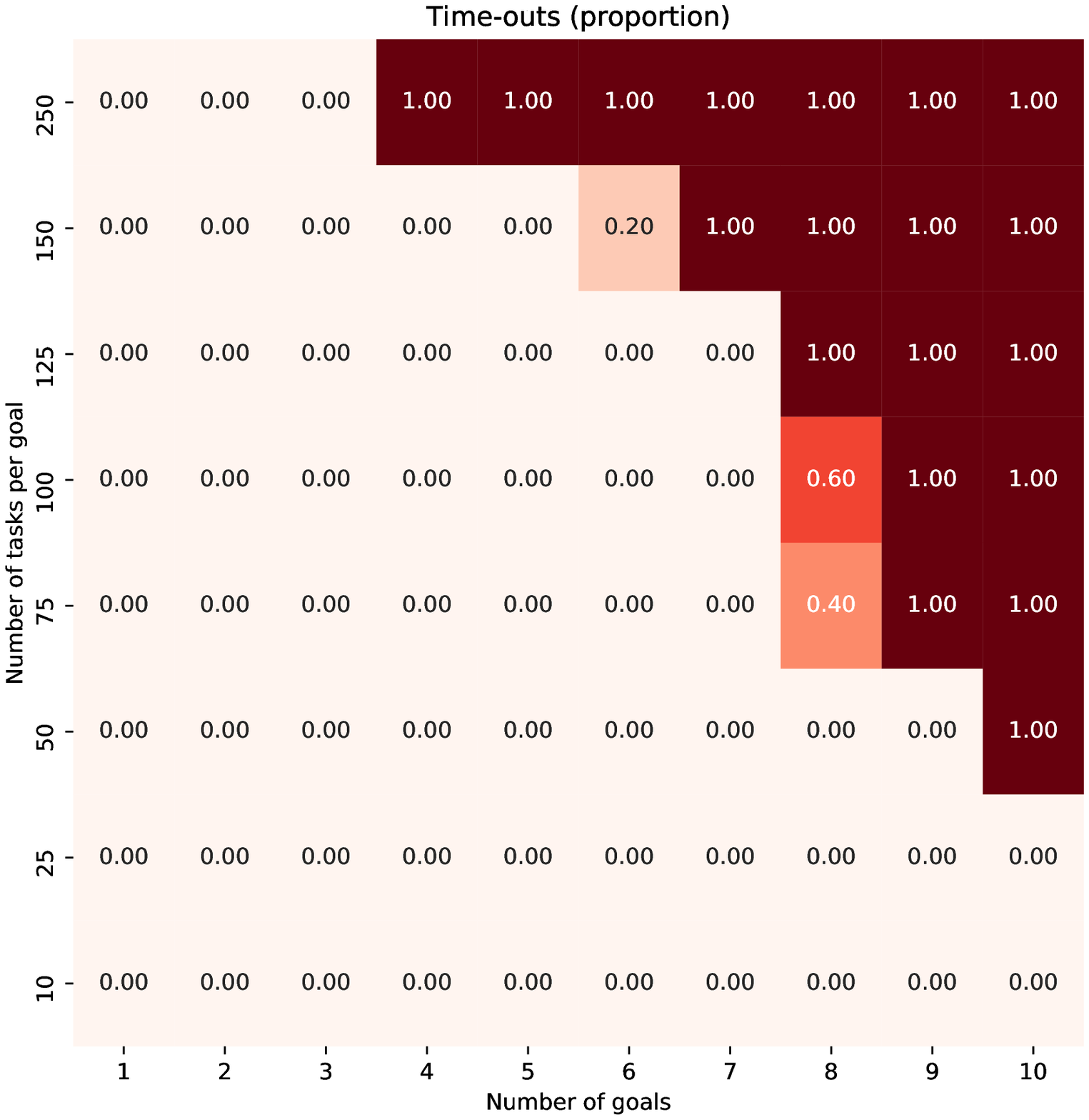}
            \caption{Reliability tests for scenario: 16 working hours, 15 minutes average task time estimate, 1 possible task duration.}
            \label{fig:1b_960tm_15te_tout}
        \end{figure}

\begin{figure}[!hp]
            \centering
            \includegraphics[width=0.8\textwidth]{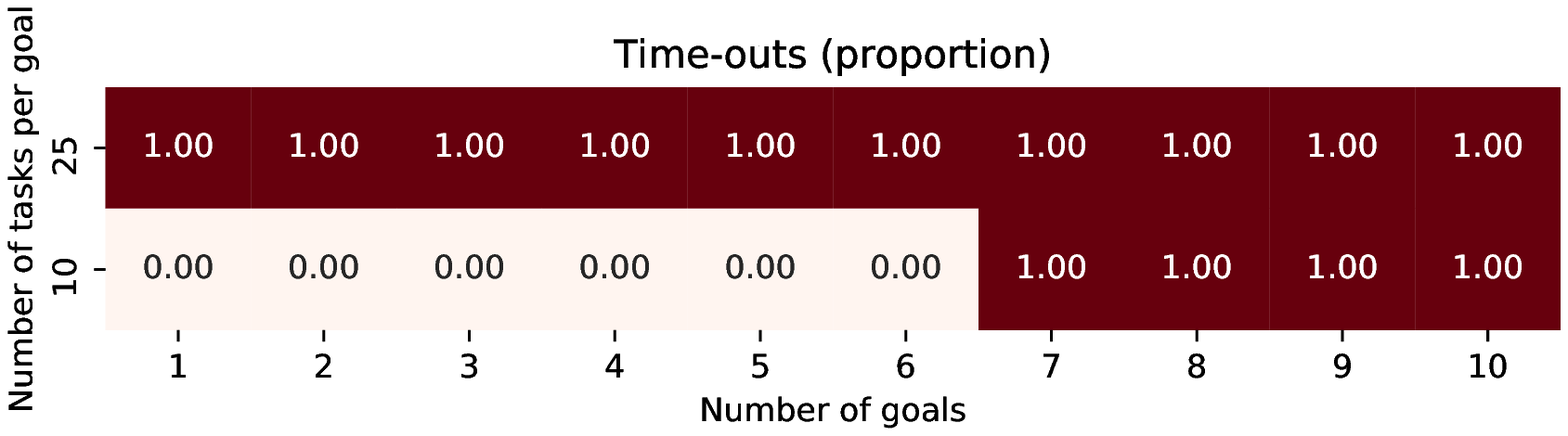}
            \caption{Reliability tests for scenario: 16 working hours, 15 minutes average task time estimate, 2 possible task durations.}
            \label{fig:2b_960tm_15te_tout}
        \end{figure}

        Results from the speed and reliability tests in the case of only 1 potential task duration for 8, 12, and 16 daily working hours are provided in Figures~\ref{fig:1b_480tm_15te_time}, \ref{fig:1b_480tm_15te_tout}, \ref{fig:1b_720tm_15te_time}, \ref{fig:1b_720tm_15te_tout}, \ref{fig:1b_960tm_15te_time}, and \ref{fig:1b_960tm_15te_tout}. According to them, we expect the API to support most of the real-world to-do lists (e.g. 5 goals wtih 150 tasks, 8 goals with 100 tasks etc.) Unfortunately, we cannot make the same statement in the case of 2 potential task durations. The results presented in Figures~\ref{fig:2b_480tm_15te_time}, \ref{fig:2b_480tm_15te_tout}, \ref{fig:2b_720tm_15te_time}, \ref{fig:2b_720tm_15te_tout}, \ref{fig:2b_960tm_15te_time}, and \ref{fig:2b_960tm_15te_tout} show that the API can support no more that 6 goals with 10 tasks per goal.
        
        In conclusion, we expect the API to be scalable for most to-do lists encountered in real-world scenarios when a single possible task duration is taken into account. Future work will be directed towards improving the scalability and reliability of the algorithm for multiple possible task durations and will report detailed information on the structure of real-world to-do lists, number of goals, number of tasks, proportion of task deadlines etc.
    
    \subsection{Comparison to non-hierarchical SMDP solving methods}
    
        The hierarchical SMDP method is optimal in cases where to-do lists have no task deadlines. However, there might exist discrepancies in cases where non-atomic goal execution is required in order to meet task deadlines. We illustrate this observation via a simple example.
        
        Let there be two goals with two tasks each.
        \begin{itemize}
            \item Goal 1 (G1) | Value: 500
            \begin{itemize}
                \item Task 1 (G1-T1) | Time estimate: 1 | Deadline: 1
                \item Task 2 (G1-T2) | Time estimate: 3 | Deadline: 6
            \end{itemize}
            \item Goal 2 (G2) | Value: 500
            \begin{itemize}
                \item Task 1 (G2-T1) | Time estimate: 2 | Deadline: 3
                \item Task 2 (G2-T2) | Time estimate: 4 | Deadline: 10
            \end{itemize}
        \end{itemize}
        The optimal solution obtained by a non-hierarchical SMDP is to execute the following sequence of actions: G1-T1 $\rightarrow$ G2-T1 $\rightarrow$ G1-T2 $\rightarrow$ G2-T2, but policy found by the hierarchical SMDP does not match the optimal solution.
        The policy obtained by the hierarchical SMDP chooses and atomically executes the most-valuable goal by trading off between the loss incurred by not attaining deadlines and the goal values.

        Theoretically, the number of mismatches in the sequence of actions in the worst case is $O(|\mathcal{G}| \cdot \tilde{n})$, where $|\mathcal{G}|$ is the total number of goals and $\tilde{n}$ is the highest number of tasks within a goal in a to-do list. This occurs when tasks have to be executed sequentially (e.g. G1-T1 $\rightarrow$ G2-T1 $\rightarrow$ ... $\rightarrow$ GN-T1 $\rightarrow$ G1-T2 $\rightarrow$ ... $\rightarrow$ GN-T2 $\rightarrow$ ...). Furthermore, it holds in general that the value difference between the hierarchical and non-hierarchical SMDP method is equivalent to the loss incurred by missing task deadlines that are not detected as attainable. However, we believe that the hierarchical decomposition of the solution method does not introduce large deviations from optimality since the task incentives are updated after every task completion. We leave investigating this issue in real-world to-do lists for future work.
        
\begin{figure}[!bp]
        \centering
        \includegraphics[width=0.8\textwidth]{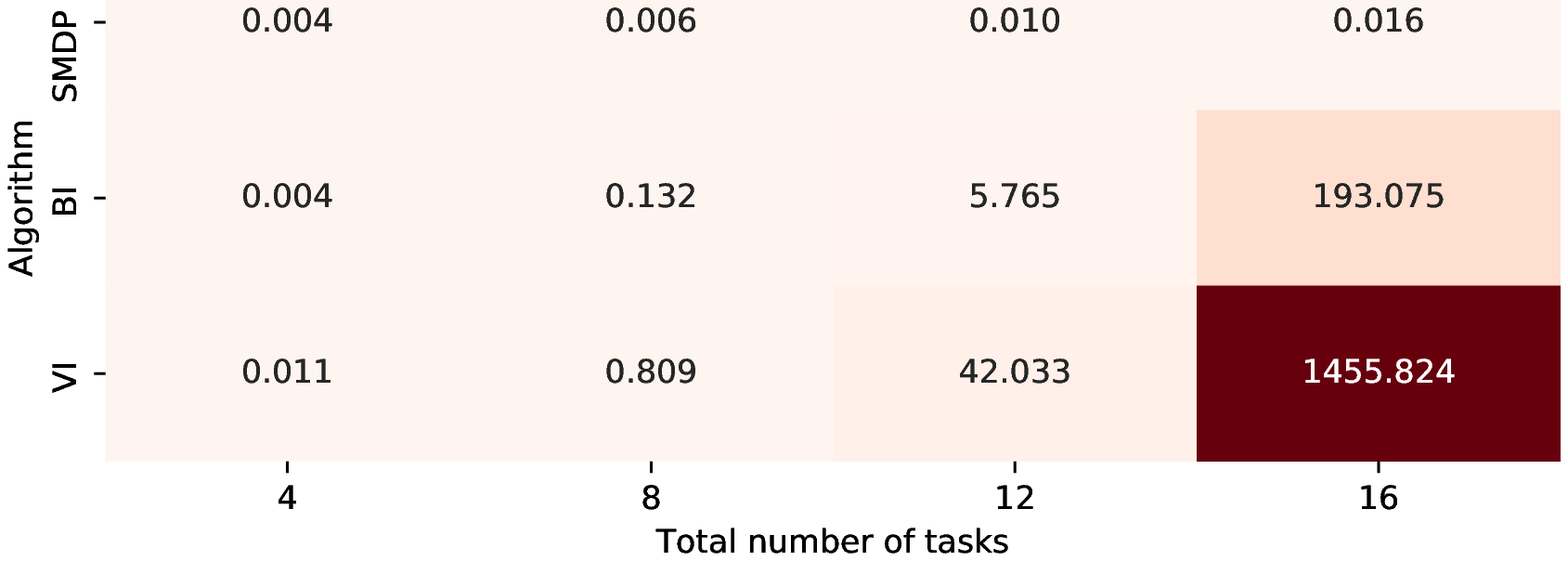}
        \caption{Speed test comparison of various solving methods (SMDP; Backward Induction - BI; Value Iteration - VI) as a function of \textit{total} number of the tasks with a single possible task duration.}
        \label{fig:bivi_hmap}
    \end{figure}

\begin{figure}[!bp]
        \centering
        \includegraphics[width=0.8\textwidth]{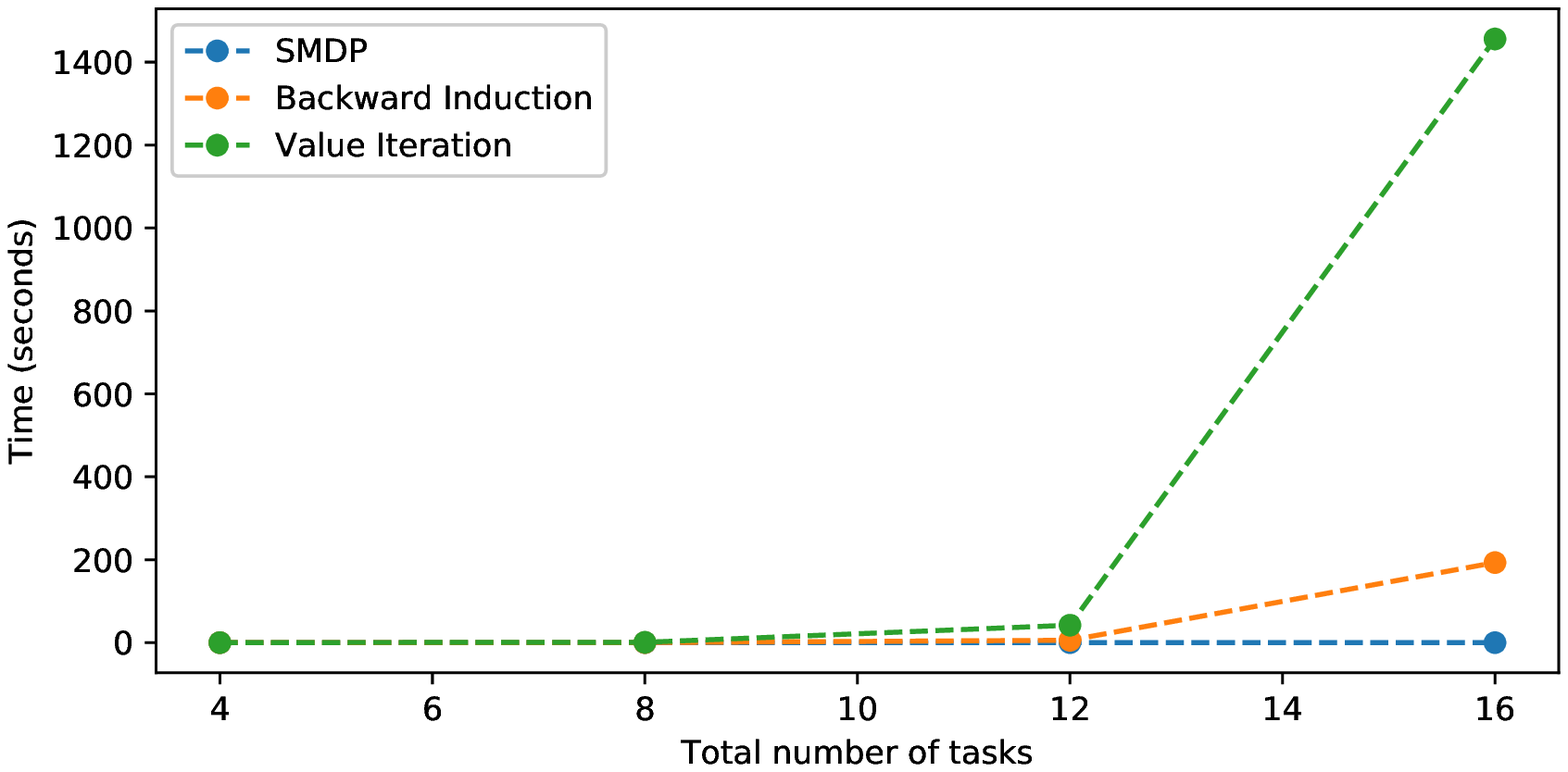}
        \caption{Speed test comparison of various solving methods (SMDP; Backward Induction - BI; Value Iteration - VI) as a function of \textit{total} number of the tasks with a single possible task duration.}
        \label{fig:bivi_lines}
    \end{figure}

        Regarding the execution time, we found that in comparison with other algorithms, the hierarchical SMDP method clearly outperforms alternative methods such as \textit{Backward Induction} (BI) and \textit{Value Iteration} (VI). A visual representation of this observation is shown in Figures~\ref{fig:bivi_hmap} and \ref{fig:bivi_lines}, where the BI and VI algorithms perform worse even for a small amount of tasks (up to 16) with a single possible task duration. Concretely, while the \textit{Backward Induction} algorithm struggles to solve a to-do list with 16 tasks (runtime of about 193 seconds), the hierarchical SMDP algorithm is able to solve 10 to-do lists with a total number of 800 tasks in comparable time (about 191 seconds).

\section{Future work}  \label{sec:future_work}

    We consider multiple potential ways to improve the API in order to make its functionality even closer to real-world demands. On one hand, usability enhancements will include supporting tasks that contribute to multiple goals simultaneously as well as multi-level hierarchical to-do lists. Furthermore, we will support dependencies between tasks imposed by users in order to form sequences of tasks that have to be executed in a particular order. Additionally, we will allow users to specify desired workload for each weekday.
    On the other hand, algorithmic enhancements will include support for multiple possible task durations in order to model multiple real-life situations while retaining (or even decreasing) computational cost. Decreasing the time complexity is of the highest priority. One possible future improvement would be to produce fast responses after making minor changes in the input information even if the changes may modify the solution.

        \begin{figure}[!hp]
            \centering

            \begin{itemize}
            
                \item[\textbf{\clrBlue{1)}}] \textbf{\clrBlue{
                    Learn mathematical foundations of machine learning (deadline: 2021-04-30; value: 1000)
                }}
                \begin{itemize}
                    \item[$\bullet$] \clrBlue{Lectures}
                    \begin{itemize}
                        \item[$\bullet$] \clrBlue{Read lecture notes for the next lecture (time est: 3 hours; ``do'' days: Wednesdays)}
                        \item[$\bullet$] \clrBlue{Attend lecture (time est: 2 hours; ``do'' days: Thursdays)}
                    \end{itemize}
                    
                    \item[$\bullet$] \clrBlue{Weekly assignments}
                    \begin{itemize}
                        \item[$\bullet$] \clrBlue{Solve exercises (time est: 3 hours; ``do'' days: Mondays)}
                        \item[$\bullet$] \clrBlue{Write down solutions in \LaTeX (time est: 1 hour; ``do'' days: Mondays)}
                        \item[$\bullet$] \clrBlue{Submit solutions (time est: 30 minutes; ``do'' days: Thursdays)}
                    \end{itemize}
                    
                    \item[$\bullet$] \clrBlue{Final exam (time est: 2 hours; ``do'' date: 2021-02-20)}
                    \item[$\bullet$] \clrBlue{Everything else (time est: 60 hours)}
                \end{itemize}
                
                \item[\textbf{\clrOlive{2)}}] \textbf{\clrOlive{
                    Take part in the seminar on causal inference (deadline: 2020-09-30; value: 500)
                }}
                \begin{itemize}
                
                    \item[$\bullet$] \clrOlive{Prepare for the next session}
                    \begin{itemize}
                        \item[$\bullet$] \clrOlive{Read Spohn’s ``Causation: An Alternative'' (time est: 4 hours; ``do'' day: Wednesday)}
                    \end{itemize}
                    
                    \item[$\bullet$] \clrOlive{Presentation}
                    \begin{itemize}
                        \item[$\bullet$] \clrOlive{Read Hájek’s ``Interpretations of Probability'' (time est: 2 hours; ``do'' day: today)}
                        \item[$\bullet$] \clrOlive{Compose slides for presentation (time est: 2 hours; deadline: 2020-09-14)}
                        \item[$\bullet$] \clrOlive{Send presentation (time est: 30 minutes; deadline: 2020-09-21)}
                        \item[$\bullet$] \clrOlive{Practice presentation (time est: 10 hours; deadline: 2020-09-28)}
                        \item[$\bullet$] \clrOlive{Presentation day (time est: 2 hours; deadline: 2020-09-28)}
                    \end{itemize}
                    
                    \item[$\bullet$] \clrOlive{Everything else (time est: 20 hours)}
                \end{itemize}
                
                \item[\textbf{\clrFuchsia{3)}}] \textbf{\clrFuchsia{Pursue a master’s degree in machine learning (deadline: 2021-09-30; value: 5000)}}
                \begin{itemize}
                    \item[$\bullet$] \clrFuchsia{Summer semester 2020}
                    \begin{itemize}
                        \item[$\bullet$] \clrFuchsia{Read exam regulations (time est: 1 hour; ``do'' day: today)}
                        \item[$\bullet$] \clrFuchsia{Register for exams (time est: 2 hours; deadline: 2020-08-30)}
                    \end{itemize}
                    
                    \item[$\bullet$] \clrFuchsia{Winter semester 2020}
                    \begin{itemize}
                        \item[$\bullet$] \clrFuchsia{Choose courses for the next semester (time est: 4 hours; deadline: 2020-10-31)}
                    \end{itemize}
                    
                    \item[$\bullet$] \clrFuchsia{Summer semester 2021}
                    \begin{itemize}
                        \item[$\bullet$] \clrFuchsia{Explore potential topics for master thesis (time est: 50 hours; deadline: 2021-03-31)}
                        \item[$\bullet$] \clrFuchsia{Write master thesis (time est: 400 hours)}
                        \item[$\bullet$] \clrFuchsia{Prepare master thesis defense (time est: 50 hours)}
                        \item[$\bullet$] \clrFuchsia{Defend master thesis (time est: 2 hours; ``do'' date: 2021-09-30)}
                    \end{itemize}
                \end{itemize}
                
                \item[+] Today's working hours: 10 hours
                \item[+] Typical day's working hours: 10 hours
            \end{itemize}
            
            \caption{Unincentivized to-do list.}
            \label{fig:use_case_input}
        \end{figure}
        
        \begin{figure}
            \centering

            % "bias": 683.2111000706025
            % "scale": 1.1,
            % "time": "2020-08-03 08:00"
    
            \begin{tabular}{l l r}
                & \textbf{To do} & \textbf{Points}
                \\
                \hline
                \\
                \textbf{Today}
                & \clrBlue{1) Solve exercises (takes about 4 hours and 11 minutes)} & \clrBlue{686} \\
                & \clrBlue{1) Write solutions in LaTeX (takes about 1 hour and 24 minutes)} & \clrBlue{684} \\
                & \clrOlive{2) Read Hájek’s “Interpretations of Probability” (takes about 2 hours and 47 minutes)} & \clrOlive{683} \\
                & \clrFuchsia{3) Read exam regulations (takes about 1 hour and 24 minutes)} & \clrFuchsia{619} \\
            \end{tabular}
            
            \vspace{0.2cm}
                \textbf{Complete} \contour{black}{$\xdownarrow{0.5cm}$} \normalsize \textbf{1\textsuperscript{st} task}
            \vspace{0.5cm}
            
            % "bias": 683.2111000706025
            % "scale": 1.1,
            % "time": "2020-08-03 08:35"

            \begin{tabular}{l l r}
                & \textbf{To do} & \textbf{Points}
                \\
                \hline
                \\
                \textbf{Today}
                & \clrBlue{\st{1) Solve exercises (takes about 4 hours and 11 minutes)}} & \\
                & \clrBlue{1) Write solutions in LaTeX (takes about 1 hour and 24 minutes)} & \clrBlue{684} \\
                & \clrOlive{2) Read Hájek’s “Interpretations of Probability” (takes about 2 hours and 47 minutes)} & \clrOlive{683} \\
                & \clrFuchsia{3) Read exam regulations (takes about 1 hour and 24 minutes)} & \clrFuchsia{619} \\
            \end{tabular}
            \caption{Incentivized to-do list generated on 2020-08-03 (Monday) at 08:00. URL parameters: \newline
            \scriptsize{\inlinecode{smdp/mdp/30/14/inf/0/inf/0/inf/false/0/max/0.999999/0.1/2/1.39/0.0001/0.01/tree/u123/getTasksForToday}}}.
            \label{fig:use_case_output}
        \end{figure}
            
        % ===== 1 bin =====
        
        % ===== 480 today minutes - 15 time estimate | time | 1 bin =====
        
        % ===== 480 today minutes - 15 time estimate | time | 2 bins =====
        
        % ===== 480 today minutes - 15 time estimate | timeouts | 1 bin =====

        % ===== 480 today minutes - 15 time estimate | timeouts | 2 bins =====
        
        % ===== 720 today minutes - 15 time estimate | time | 1 bin =====
        
        % ===== 720 today minutes - 15 time estimate | time | 2 bins =====
        
        % ===== 720 today minutes - 15 time estimate | timeouts | 1 bin =====
        
        % ===== 720 today minutes - 15 time estimate | timeouts | 2 bins =====
        
        % ===== 960 today minutes - 15 time estimate | time | 1 bin =====
        
        % ===== 960 today minutes - 15 time estimate | time | 2 bins =====
        
        % ===== 960 today minutes - 15 time estimate | timeouts | 1 bin =====
        
        % ===== 960 today minutes - 15 time estimate | timeouts | 2 bins =====

    % ===== BI and VI | heatmap =====

    % ===== BI and VI | plot =====

\section*{Acknowledgment}
This work was supported by grant number 1757269 from the National Science Foundation.

\newpage

\addcontentsline{toc}{section}{References}
\bibliographystyle{plain}
\bibliography{biblio.bib}
\newpage

\addcontentsline{toc}{section}{Appendix}
\section*{Appendix}  \label{sec:appendix}
 
    \renewcommand{\thesubsection}{A\arabic{subsection}}

    \subsection{API input}  \label{sec:api_input}
        The API receives a POST request with a specific URL and body in JSON format as input from the gamification application. The general pattern of the URL is as follows: \newline
        \small{\inlinecode{http://<server>/api/<compulsoryParameters>/<additionalParameters>/tree/<userID>/<functionName>}}\normalsize.
        \\
        \begin{itemize}
            \item \inlinecode{<server>}: DNS or IP address of the server. 
            \item \inlinecode{<compulsoryParameters>}: Parameters that are \textit{independent} of the incentivizing method.
            \item \inlinecode{<additionalParameters>}: Parameters that are related to the incentivizing method.
            \item \inlinecode{<userID>}: A unique user identification code.
            \item \inlinecode{<functionName>}: The functionality that the API should provide.
        \end{itemize}
        Please note that you must set up a MongoDB database in order to be able to store the information generated by the API, and to add the parameter as a configuration variable so that it can be accessed in the main block in the file \inlinecode{app.py}. For details on each of these items, please refer to the \href{https://github.com/RationalityEnhancement/todolistAPI/blob/master/README.md}{\texttt{README.md}} file of the repository.
        
        Additionally, the request must contain a body in JSON format with the following information:
        \begin{itemize}
            \item \inlinecode{currentIntentionsList}: List of tasks that have already been scheduled. Each item in this list represents a scheduled task and it has to contain the following information:
            \begin{itemize}
                \item \inlinecode{\_c}: Goal code/number.
                \item \inlinecode{\_id}: Unique identification code of the scheduled task.
                \item \inlinecode{d}: Whether the scheduled task has been completed or not.
                \item \inlinecode{nvm}: Whether the scheduled task has been marked to be completed at some other time.
                \item \inlinecode{t}: Title of the scheduled task.
                \item \inlinecode{vd}: Value of the scheduled task.
            \end{itemize}
    
            \item \inlinecode{projects}: Tree of goals and their respective tasks. Each item (goal or task) is composed of the following information:
            \begin{itemize}
                \item \inlinecode{id}: Unique identification code of the item.
                \item \inlinecode{nm}: Title of the item.
                \item \inlinecode{lm}: Time stamp of item's last modification.
                \item \inlinecode{cp}: Time stamp of item's completion.
                \item \inlinecode{ch}: Sub-items of the current item.
            \end{itemize}
            
            \item \inlinecode{timezoneOffsetMinutes}: Time difference in minutes between user's time zone and UTC.
            \item \inlinecode{today\_hours}: Number of hours that a user would like to work on the current day (today).
            \item \inlinecode{typical\_hours}: Number of hours that a user would like to work on a typical day.
            \item \inlinecode{userkey}: Unique identification code of the user.
            \item \inlinecode{updated}: Time stamp of the last modification of the items in the \inlinecode{projects} tree.
        \end{itemize}
        
        Each to-do-list item (i.e. goal or task) title follows patterns that encode all the necessary information. The following list describes these patterns in detail:
        \begin{itemize}
            \item \inlinecode{\#CG<N>\_<goal\_name>} defines a goal name, where \inlinecode{<N>} is the number of the goal and \inlinecode{<goal\_name>} is the actual goal name specified by the user.
            \item \inlinecode{==<value>} defines a value of a goal/tasks, where \inlinecode{<value>} $\in \mathbb{Z}_{\ge 0}$.
            \item \inlinecode{DUE:<YYYY-MM-DD> <HH:mm>} defines a deadline, where \inlinecode{<YYYY-MM-DD>} defines a date according to the ISO format and \inlinecode{<HH:mm>} defines a 24-hours day time. If \inlinecode{<HH:mm>} is not provided, then 23:59 is taken as a default day-time value.
            \item \inlinecode{$\sim\sim$<time\_estimate> <time\_unit>} defines a time estimate for a task, where \inlinecode{<time\_estimate> min} $\in \mathbb{N}$ corresponds to the number of minutes \textit{or} \inlinecode{<time\_estimate> h} $\in \mathbb{R}_{> 0}$ corresponds to the amount of hours.
            \item \inlinecode{\#HOURS\_TYPICAL ==<hours>} defines the total number of hours per day, i.e. the amount of hours $\in (0, 24]$ that a user wants to work on a typical day.
            \item \inlinecode{\#HOURS\_TODAY ==<hours>} defines the total number of hours for today, i.e. the amount of hours $\in (0, 24]$ that a user wants to work today.
            \item Scheduling tags that users can accompany to their tasks:
            \begin{itemize}
                \item \inlinecode{\#daily} represents a task that is repetitive on a daily basis.
                \item \inlinecode{\#future} represents a task that has to be scheduled at some point in the future, but not at the moment.
                \item \inlinecode{\#today} represents a task that has to be scheduled today.
                \item \inlinecode{\#<weekday>} represents a task that has to be scheduled on a specific weekday (where \inlinecode{weekday} is a day from Monday to Sunday). If this task is repetitive on a weekly basis, a plural suffix is appended to the same tag, i.e. \inlinecode{\#<weekday>s}.
                \item \inlinecode{\#weekdays} represents a repetitive task that has to be scheduled on each working day (from Monday to Friday).
                \item \inlinecode{\#weekends} represents a repetitive task that has to be scheduled during weekends (Saturday and Sunday).
                \item \inlinecode{\#YYYY-MM-DD} represents a task that has to be scheduled on a specific day according to the ISO standard (year-month-day).
            \end{itemize}
        \end{itemize}
    
    \subsection{API output}  \label{sec:api_output}
        After generating incentives for each task in a to-do-list, the API selects a subset of them and it proposes an incentivized daily schedule as output. The output is a list of dictionaries in JSON format and it contains the following information for each task in the list:
        \begin{itemize}
            \item \inlinecode{id}: Unique identification code of the task.
            \item \inlinecode{nm}: Human-readable name of the task.
            \item \inlinecode{lm}: Time stamp of task's last modification.
            \item \inlinecode{est}: Time estimate of the task.
            \item \inlinecode{parentId}: Unique identification code of the goal which the task belongs to.
            \item \inlinecode{pcp}: Whether the parent node (i.e. goal) has been completed.
            \item \inlinecode{val}: Generated incentive for the task.
        \end{itemize}

\end{document}